\documentclass[letterpaper]{article} 
\usepackage{aaai25}  
\usepackage{times}  
\usepackage{helvet}  
\usepackage{courier}  
\usepackage[hyphens]{url}  
\usepackage{graphicx} 
\usepackage{natbib}  
\usepackage{caption} 
\usepackage{algorithm}
\usepackage{algorithmic}

\usepackage{amsthm}
\usepackage{amsmath}
\usepackage{amssymb}
\usepackage{booktabs}
\usepackage{multirow}
\usepackage{bbding}
\usepackage[table]{xcolor}
\usepackage{color}
\usepackage{xspace}
\def\onedot{\ifx\@let@token.\else.\null\fi\xspace}
\def\eg{\emph{e.g}\onedot}
\def\Vec#1{{\boldsymbol{#1}}}
\def\Mat#1{{\boldsymbol{#1}}}
\usepackage{newfloat}
\usepackage{listings}
\DeclareCaptionStyle{ruled}{labelfont=normalfont,labelsep=colon,strut=off} 
\floatstyle{ruled}
\newfloat{listing}{tb}{lst}{}
\floatname{listing}{Listing}
\title{SVasP: Self-Versatility Adversarial Style Perturbation \\ for Cross-Domain Few-Shot Learning}
\author{
 Wenqian Li\textsuperscript{\rm 1,2}, Pengfei Fang\textsuperscript{\rm 1,2}$^*$, Hui Xue\textsuperscript{\rm 1,2}$^*$
}
\affiliations{
    \textsuperscript{\rm 1}School of Computer Science and Engineering, Southeast University, Nanjing 210096, China\\
    \textsuperscript{\rm 2}Key Laboratory of New Generation Artificial Intelligence Technology and Its Interdisciplinary Applications (Southeast University), Ministry of Education, China\\
    \{wenqianli.li, fangpengfei, hxue\}@seu.edu.cn

}

\usepackage{bibentry}

\begin{document}

\maketitle

%
\begin{abstract}
Cross-Domain Few-Shot Learning (CD-FSL) aims to transfer knowledge from seen source domains to unseen target domains, which is crucial for evaluating the generalization and robustness of models. Recent studies focus on utilizing visual styles to bridge the domain gap between different domains. However, the serious dilemma of gradient instability and local optimization problem occurs in those style-based CD-FSL methods. This paper addresses these issues and proposes a novel crop-global style perturbation method, called \underline{\textbf{S}}elf-\underline{\textbf{V}}ersatility \underline{\textbf{A}}dversarial \underline{\textbf{S}}tyle \underline{\textbf{P}}erturbation (\textbf{SVasP}), which enhances the gradient stability and escapes from poor sharp minima jointly. Specifically, SVasP simulates more diverse potential target domain adversarial styles via diversifying input patterns and aggregating localized crop style gradients, to serve as global style perturbation stabilizers within one image, a concept we refer to as self-versatility. Then a novel objective function is proposed to maximize visual discrepancy while maintaining semantic consistency between global, crop, and adversarial features. Having the stabilized global style perturbation in the training phase, one can obtain a flattened minima in the loss landscape, boosting the transferability of the model to the target domains. Extensive experiments on multiple benchmark datasets demonstrate that our method significantly outperforms existing state-of-the-art methods. Our codes are available at https://github.com/liwenqianSEU/SVasP.
\end{abstract}

\section{Introduction}\label{sec:intro}
Deep learning models have achieved significant advancements in visual recognition when trained with abundant labeled samples. However, in many real-world applications, such as rare disease diagnosis, large training datasets with reliable annotations are not always feasible. To address this limitation, Few-Shot Learning (FSL) methods have been developed to enable models to generalize to novel classes with only a few samples per class~\cite{snell2017prototypical, sung2018learning, feng2024wave}. In addition to the challenge of limited data, there is often a domain gap between the source domains and target domains in practical scenarios, which presents a critical challenge. Consequently, Cross-Domain Few-Shot Learning (CD-FSL) methods have been explored to transfer domain-agnostic knowledge from multiple well-annotated source domains to target domains with limited labeled data~\cite{tseng2020cross, guo2020broader, triantafillou2019meta, feng2024transferring}. Among the various CD-FSL settings, Single Source CD-FSL addresses domain shifts more realistically by restricting the model to access only one source domain during training.

\begin{figure}[t!]
\centering
\includegraphics[width=0.49\textwidth]{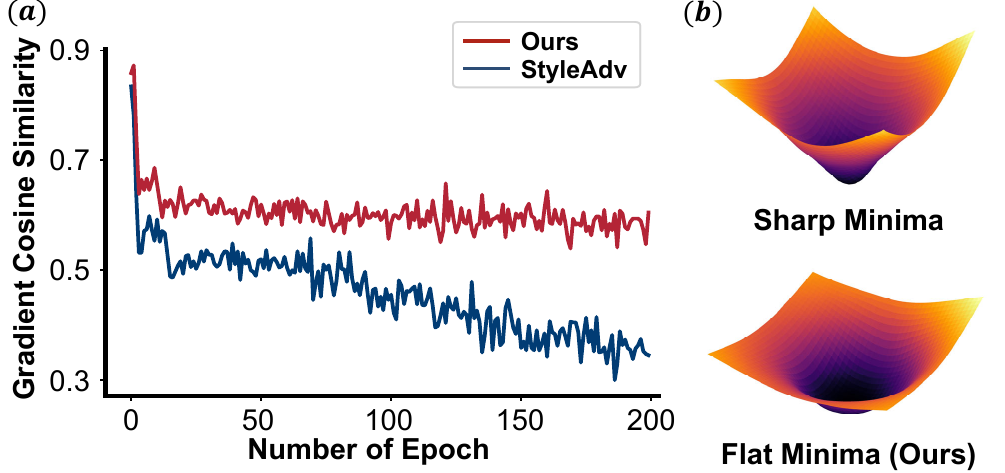} 
\caption{SVasP stabilizes the gradients and escapes from poor sharp minima. (a) demonstrates the gradient cosine similarity between epochs for displaying ground representations of gradient stability, and the larger the cosine similarity, the more stable gradient update direction. (b) demonstrates the proposed approach ensures that the model converges to a flat minima and is robust to domain shifts.}
\label{fig:motivation}
\end{figure}

Recent studies have explored perturbing the styles of source domain images to facilitate models acquiring more domain-agnostic knowledge from a single source domain~\cite{kim2024randomized, zhang2022free, wang2022remember, zhong2022adversarial}. By recognizing image styles (\eg, mean and standard deviation) as key domain characteristics~\cite{zhou2020domain}, these studies aim to enhance model generalization and mitigate domain shifts by altering these domain-specific attributes~\cite{feng2023genes, xie2024kind}. Although style-based methods have demonstrated effectiveness in Cross-Domain Few-Shot Learning (CD-FSL), they remain suboptimal due to the inherent differences between domains and the varied optimization paths for adversarial perturbations. As a result, models may overfit to noise or specific samples, becoming overly dependent on the source domain and thereby limiting their generalization capabilities.

Recently, StyleAdv~\cite{fu2023styleadv} has addressed domain shifts by augmenting the original styles with signed style gradients. Although effective in CD-FSL tasks, StyleAdv exhibits significant gradient instability. As illustrated in Figure \ref{fig:motivation} (a), we measure the gradient cosine similarity between the forward and backward gradients to assess gradient stability. However, we observe a continuous decline and severe oscillations in gradient cosine similarity, indicating that stable gradient optimization is unattainable. This instability is attributed to the absence of target domains and inadequate collection of gradient information from the source domain, which can misdirect adversarial style attacks~\cite{wang2021enhancing}. Moreover, StyleAdv's strategy of employing minimal perturbations for adversarial training tends to make the model overly sensitive to such perturbations, thereby undermining its robustness.

To address these challenges, we leverage diverse inputs from a single source domain to enhance style diversity and propose a novel framework called Self-Versatility Adversarial Style Perturbation (\textbf{SVasP}). We argue that localized crop style gradients play a crucial role in model performance. The core idea of SVasP is to enhance the transferability of source domain knowledge by integrating localized crop style gradients with global style optimization. Contrary to StyleAdv, our SVasP improves the stability of model gradient in the optimization phase, shown in Figure \ref{fig:motivation} (a). As training progresses, the issue of gradient oscillation is effectively mitigated, allowing the model to escape sharp minima and achieve smoother, flatter minima, which are more conducive to improving the model’s generalization, as illustrated in Figure \ref{fig:motivation} (b).

Specifically, our method employs a structure of inner and outer iterations. In each outer iteration, sections of the benign image are randomly cropped and resized for use in the subsequent inner iterations. During each inner iteration, we iteratively generate and integrate all crop style gradients, applying them to target the global style of the benign image. The central concept of our approach is to stabilize the gradients by incorporating as much relevant gradient information from the source domain as possible. To the best of our knowledge, this is the inaugural study exploring the impact of localized style gradients on model generalization.

The main contribution of our paper is three-fold: 
\begin{itemize}
    \item We propose a new framework called SVasP that incorporates crop style gradients with the global style gradients within a image itself, which is called self-versatility, to efficiently stabilize gradients for adversarial style attack and escape from the sharp minima.
    \item We design a novel objective function, named Discrepancy \& Consistency Optimization (DCO) to maximize visual discrepancy between seen and unseen domains while maintaining semantic consistency.
    \item  We conduct extensive experiments on multiple benchmark datasets and validate the effectiveness of our modules. The quantitative results show that our proposed SVasP significantly improves the model's generalizability over other state-of-the-art(SOTA) methods.
\end{itemize}

\section{Related Work} \label{sec:related}
\subsubsection{Cross-Domain Few-Shot Learning.}
Cross-Domain Few-Shot Learning (CD-FSL) aims to train a model on source domains that can effectively generalize to target domains, first introduced in \cite{schen2019closer}. Key benchmarks include BSCD-FSL \cite{guo2020broader}, \textit{mini}-CUB~\cite{tseng2020cross}, and Meta-Dataset~\cite{triantafillou2019meta}.

CD-FSL methods can be categorized based on access to target domain data: Single Source CD-FSL~\cite{zou2024flatten, sun2021explanation, wang2021cross, hu2022adversarial}, unlabeled target-domain CD-FSL~\cite{islam2021dynamic, phoo2020self, zheng2022cross}, and labeled target-domain CD-FSL~\cite{fu2021meta, zhuo2022tgdm, fu2022me}. 
This paper focuses on the most realistic and challenging setting, Single Source CD-FSL, where only a source domain dataset is accessible.

\subsubsection{Input Diversity for Domain Shift.} 
To address domain shift, many methods enhance input diversity. In domain generalization, MiRe~\cite{chen2022mix} mixes images from different domains, and CreTok~\cite{feng2024redefining} combines tokens for creative generation. In object detection, DoubleAUG~\cite{qi2024doubleaug} exchanges RGB channels, and RECODE~\cite{li2024zero} decomposes visual features into subject, object, and spatial features. In CD-FSL, LDP-net~\cite{zhou2023revisiting} extracts local features, TGDM~\cite{zhuo2022tgdm} and meta-FDMixup~\cite{fu2021meta} mix source and auxiliary data, and ConFeSS~\cite{das2021confess} and~\cite{zheng2022cross} use different augmentation methods. These augmentation methods generate diverse input patterns and more generic features for transfer. However, none of these works consider the gradient instability problem, which is a critical issue in Single Source CD-FSL.

\subsubsection{Gradient-based Optimization.} 
Various gradient-based optimization methods improve model robustness and generalization. GradNorm~\cite{chen2018gradnorm} and GAM~\cite{zhang2023gradient} explore gradient normalization techniques. CGDM~\cite{du2021cross} minimizes the discrepancy between gradients from source and target samples. Fishr~\cite{rame2022fishr} aligns domain-level loss landscapes by leveraging gradient covariances, and PCGrad~\cite{yu2020gradient} addresses conflicting gradients in multi-task learning. 
However, these methods often overlook diverse patterns, such as crop image style gradients, which limits their effectiveness in addressing model overfitting.

\begin{figure*}[t!]
\centering
\includegraphics[width=1\textwidth]{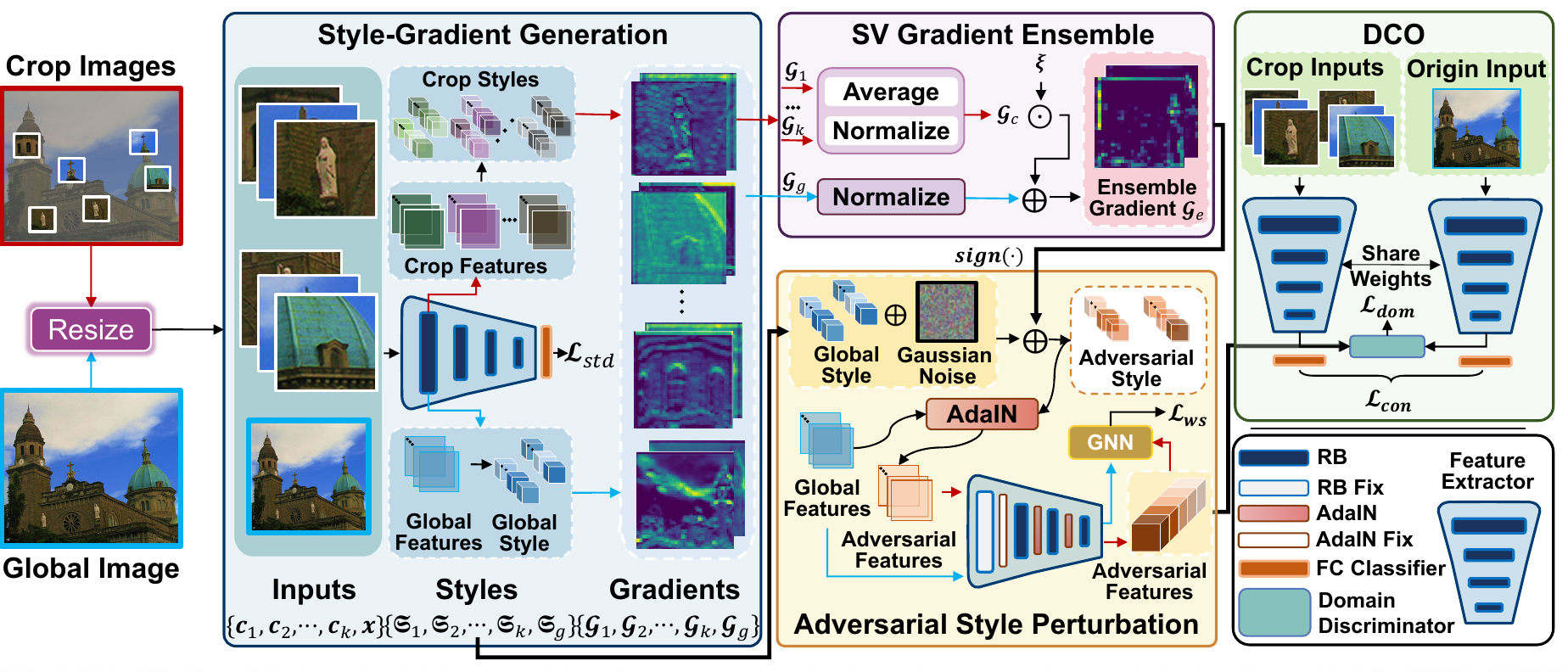} %
\caption{Overview of our proposed methods SVasP. ``RB'' is an abbreviation for ResNet Block. Random cropping the benign image and generates several crop images. Then, four main modules are performed: a) Generate the gradients of both crop and global styles (illustration with $B_1$); b) Integrate localized crop style gradients into the global style gradients; c) Perform adversarial style perturbation based on AdaIN method; d) DCO: Maximize domain visual discrepancy and global-crop consistency.}
\label{fig:structure}
\end{figure*}
\section{Methodology}
This section introduces the proposed novel framework \textbf{SVasP}, designed for CD-FSL. An overview of our method is depicted in Figure \ref{fig:structure}.
\subsection{Problem Formulation}
    We focus on the Single Source CD-FSL setting where only a source dataset $\mathcal{D}^s$ can be accessed while the target dataset $\mathcal{D}^t$ is forbidden. Notably, for CD-FSL, $C(\mathcal{D}^s) \cap C(\mathcal{D}^t) = \emptyset$, $P(\mathcal{D}^s) \neq P(\mathcal{D}^t)$, where $C(\cdot)$ and $P(\cdot)$ denote the categories and distributions of the source and target dataset, respectively. Moreover, episode training is used in this work. Specifically, to simulate the $N$-way $K$-shot problem, $N$ classes are selected and $K$ samples per class are chosen to form the support set $\mathcal{S} = \{\Mat{x}_i^s, {y}_i^s\}_{i=1}^{n_s}$, where $n_s = NK$. And the same $N$ classes with another $M$ images are used to construct the query set $\mathcal{Q} = \{\Mat{x}_i^q\}_{i=1}^{n_q}$, where $n_q = NM$. Therefore, an episode $\mathcal{T} = (\mathcal{S}, \mathcal{Q})$ is constituded, comprising of a support set $\mathcal{S}$ and a query set $\mathcal{Q}$, and $|\mathcal{T}| = N (K + M)$. The goal is to classify the images of the query set by training a feature extractor and a classification head on the support set.
\subsection{SVasP}
\subsubsection{Overview.} The proposed model contains a CNN/ViT backbone $E$, a domain discriminator $f_{dom}$, a global FC classifier $f_{g}$ and a FSL relation classifier $f_{re}$ with learnable parameter $\theta_E$, $\theta_{dom}$, $\theta_{g}$ and $\theta_{re}$, respectively. 

The network consists of four components: Style-Gradient Generation module to produce global and crop style gradients, Self-Versatility (SV) Gradient Ensemble module to integrate the localized crop style gradients as the global perturbation stabilizers, Adversarial Style Perturbation module to simulate diverse unseen styles, and Discrepancy \& Consistency Optimization (DCO) to maximize the discrepancy between seen and unseen domains and maintains global-crop semantic consistency. 

Without accessing auxiliary data, SVasP moderates the gradient instability and achieves a flatter minima, robustly improving the model's generalizability. Further details are provided in the following sections. 

\subsubsection{Style-Gradient Generation.} In this paper, the styles of global and crop features are modeled as Gaussian distributions ~\cite{liuncertainty} and learnable parameters which will be updated by adversarial training. Specifically, for feature maps $\Mat{F} \in \mathbb{R}^{B \times C \times H \times W}$, where $B$, $C$, $H$ and $W$ denote the batch size, channel, height, width of the feature maps $\Mat{F}$, the specific formula for calculating the style $\mathfrak{S} = \{\Mat{\mu}, \Mat{\sigma}\}$ is:
\begin{equation}\label{eq:mu}
\Mat{\mu}=\frac{1}{HW}\sum_{h=1}^{H}\sum_{w=1}^{W}\Mat{F}_{B,C,h,w} ,
\end{equation}
\begin{equation}\label{eq:sigma}
\negthickspace \negthickspace \Mat{\sigma}=\sqrt{\frac{1}{HW}\sum_{h=1}^{H}\sum_{w=1}^{W}(\Mat{F}_{B,C,h,w}-\Mat{\mu})^{2} + \epsilon},
\end{equation}
where $\epsilon$ is a small value to avoid division by zero. 

Unlike directly perturbing the global style, we consider incorporating crop style gradients to stabilize the global style gradients. For each benign image and label pair $(\Mat{x}, {y})$, we randomly crop $k$ local images by the scale parameter $s = \{s_l, s_h\}$, where $s_l$ and $s_h$ denote the lower and upper bound for the area of the random cropped images respectively and get the input set $\mathbb{I} = \{\Mat{c}_1, \Mat{c}_2, \cdot\cdot\cdot, \Mat{c}_k, \Mat{x}\}$. Instead of generating the adversarial style of all blocks' features at once, we use an iterative approach. Concretely, the embedding module $E$ has four blocks $B_1$, $B_2$, $B_3$, $B_4$, and style transformation only performs on the first three blocks, as the shallow blocks produce more migratory features. For each block $B_j$ of the backbone $E$, we obtain the crop and global feature maps $\mathbb{F}^j = \{\Mat{F}_{1}^j,\Mat{F}_{2}^j, \cdot\cdot\cdot, \Mat{F}_{k}^j, \Mat{F}_g^j\}$. For each $\Mat{F}^j \in \mathbb{F}^j$, $\Mat{F}^j \in \mathbb{R}^{B \times C \times H \times W}$, $\Mat{F}^j$ is accumulated from block 1 to block $j-1$:
\begin{equation}
    \Mat{F}^j = \mathfrak{T}_j(\mathfrak{T}_{j-1}(\cdot\cdot\cdot(\mathfrak{T}_1({\Mat{I}, \mathfrak{S}_{adv}^1}), \cdot\cdot\cdot), \mathfrak{S}_{adv}^{j-1}), \mathfrak{S}_{adv}^j)
\end{equation}
where transferring features between block $j-1$ and block $j$ is formulated as:
\begin{equation}
    \mathfrak{T}_j({\Mat{F}^{j}, \mathfrak{S}_{adv}^j}) = \frac{B_j(\Mat{F}^{j-1}) - \Mat{\mu}_{F^j}}{\Mat{\sigma}_{F^j}} * \Mat{\sigma}_{adv}^j + \Mat{\mu}_{adv}^j
\end{equation}
and the style $\mathfrak{S}_{F^j} = \{\Mat{\mu}_{F^j}, \Mat{\sigma}_{F^j}\}$ of $\Mat{F}^{j}$ is caculated by Eq.~\eqref{eq:mu} and \eqref{eq:sigma}.
Thus, we can get the styles of the feature maps of $B_j$ to form the style set $\mathbb{S} = \{ \mathfrak{S}_1^j, \mathfrak{S}_2^j, \cdot\cdot\cdot, \mathfrak{S}_k^j, \mathfrak{S}_g^j\}$. Then, we continue to pass $\Mat{F}^j$ to the remainder of the backbone and the global FC classifier without performing any other operations and get the final prediction $\Vec{p}=f_{g}(B_4(\cdot\cdot\cdot (B_{j+1}(\Mat{F}^j)));\theta_{g}), \Vec{p}\in \mathbb{R}^{B \times N_c}$, where $N_c$ denotes the total number of classes. Thus the total prediction set is $\mathbb{P} = \{\Vec{p}_{1}, \Vec{p}_{2}, \cdot\cdot\cdot, \Vec{p}_{k}, \Vec{p}_{g}\}$. Therefore the classification loss can be written as:
\begin{equation}
    \mathcal{L}_{cls} = \mathcal{L}_{CE}(\Vec{p}_{g}, {y}) + \sum\nolimits_{i=1}^k\mathcal{L}_{CE}(\Vec{p}_{i}, {y}) 
\end{equation}
where $\mathcal{L}_{CE}(\cdot, \cdot)$ denotes the cross-entropy loss. 

The sequel will compute the adversarial style of block $B_j$, we omit the subscript $j$ for readability and calculate the gradients of the mean $\Mat{\mu}$ and the std $\Mat{\sigma}$ by loss back propagation: 
\begin{equation}
\begin{aligned}
\mathbb{G}^{\mu} & = \{\boldsymbol{\mathcal{G}}^{\mu}_{1}, \boldsymbol{\mathcal{G}}^{\mu}_{2}, \cdot\cdot\cdot, \boldsymbol{\mathcal{G}}^{\mu}_{k}, \boldsymbol{\mathcal{G}}^{\mu}_{g}\} \\
        & = \{\nabla_{\Mat{\mu}_{1}}\mathcal{L}_{cls}, \nabla_{\Mat{\mu}_{2}}\mathcal{L}_{cls}, \cdot\cdot\cdot, \nabla_{\Mat{\mu}_{k}}\mathcal{L}_{cls}, \nabla_{\Mat{\mu}_{g}}\mathcal{L}_{cls}\}
\end{aligned}
\label{eq:g_mu}
\end{equation}
\begin{equation}
\begin{aligned}
\mathbb{G}^{\sigma} & = \{\boldsymbol{\mathcal{G}}^{\sigma}_{1}, \boldsymbol{\mathcal{G}}^{\sigma}_{2}, \cdot\cdot\cdot, \boldsymbol{\mathcal{G}}^{\sigma}_{k}, \boldsymbol{\mathcal{G}}^{\sigma}_{g}\} \\
        & = \{\nabla_{\Mat{\sigma}_{1}}\mathcal{L}_{cls}, \nabla_{\Mat{\sigma}_{2}}\mathcal{L}_{cls}, \cdot\cdot\cdot, \nabla_{\Mat{\sigma}_{k}}\mathcal{L}_{cls}, \nabla_{\Mat{\sigma}_{g}}\mathcal{L}_{cls}\}
\end{aligned}
\label{eq:g_sigma}
\end{equation}
Other blocks' style gradients can be generated likewise. 
\subsubsection{SV Gradient Ensemble.}
Self-Versatility (SV) Gradient Ensemble module serves as the core part of our work, dedicated to bootstrapping global style gradients by integrating localized crop style gradients. We first average and normalize the style gradients of the crops to get the aggregate crop style gradients $\mathbb{G}^c = \{\boldsymbol{\mathcal{G}}_{c}^\mu,\boldsymbol{\mathcal{G}}_{c}^\sigma\}$, where: 
\begin{equation}
    \boldsymbol{\mathcal{G}}_{c}^\mu = Norm(\frac{1}{k} \sum(\boldsymbol{\mathcal{G}}^{\mu}_{1} + \boldsymbol{\mathcal{G}}^{\mu}_{2} + \cdot\cdot\cdot + \boldsymbol{\mathcal{G}}^{\mu}_{k}))
\label{eq:g_crop_mu}
\end{equation}
\begin{equation}
    \boldsymbol{\mathcal{G}}_{c}^\sigma = Norm(\frac{1}{k} \sum(\boldsymbol{\mathcal{G}}^{\sigma}_{1} + \boldsymbol{\mathcal{G}}^{\sigma}_{2} + \cdot\cdot\cdot + \boldsymbol{\mathcal{G}}^{\sigma}_{k}))
\label{eq:g_crop_sigma}
\end{equation}
Subsequently, a decay factor $\xi$ is introduced to finally get the ensemble style gradients $\mathbb{G}^e = \{\boldsymbol{\mathcal{G}}_{e}^\mu,\boldsymbol{\mathcal{G}}_{e}^\sigma\}$, where:
\begin{equation}
    \boldsymbol{\mathcal{G}}_{e}^\mu = Norm(\boldsymbol{\mathcal{G}}_g^{\mu}) + \xi \odot \boldsymbol{\mathcal{G}}_{c}^\mu
\label{eq:g_e_mu}
\end{equation}
\begin{equation}
    \boldsymbol{\mathcal{G}}_{e}^\sigma = Norm(\boldsymbol{\mathcal{G}}_g^{\sigma}) + \xi \odot \boldsymbol{\mathcal{G}}_{c}^\sigma
\label{eq:g_e_sigma}
\end{equation}
\subsubsection{Adversarial Style Perturbation.} We get the random initialized global styles $\Mat{\mathfrak{S}}_{init} = \{\Mat{\mu}_{init}, \Mat{\sigma}_{init}\}$ by adding Gaussian noise $\mathcal{N}(0,I)$, where:
\begin{equation}
    \Mat{\mu}_{init} = \Mat{\mu}_g + \varepsilon \cdot \mathcal{N}(0,I)
\label{eq:mu_init}
\end{equation}
\begin{equation}
    \Mat{\sigma}_{init} = \Mat{\sigma}_g + \varepsilon \cdot \mathcal{N}(0,I)
\label{eq:std_init}
\end{equation}
where $\varepsilon$ is set to $\frac{16}{255}$. Then, the ensemble gradients are incorporated into the initialized style to get the adversarial styles $\mathfrak{S}_{adv} = \{\Mat{\mu}_{adv}, \Mat{\sigma}_{adv} \}$, where:
\begin{equation}
    \Mat{\mu}_{adv} = \Mat{\mu}_{init} + \kappa_1 \cdot sign(\boldsymbol{\mathcal{G}}_{e}^\mu)
\label{eq:mu_adv}
\end{equation}
\begin{equation}
    \Mat{\sigma}_{adv} = \Mat{\sigma}_{init} + \kappa_2 \cdot sign(\boldsymbol{\mathcal{G}}_{e}^\sigma)
\label{eq:std_adv}
\end{equation}
Notably, $\kappa_1$ and $\kappa_2$ are chosen randomly from a given set of coefficients, which will not force a consistent change in the degree of the perturbation of $\Mat{\mu}$ and $\Mat{\sigma}$, making the model generate a more diverse range of styles. 
After obtaining the adversarial styles, style migration is performed with AdaIN method to enhance the generalizability:
\begin{equation}
    \Mat{F}_{adv} = \frac{\Mat{F}_g - \Mat{\mu}_g}{\Mat{\sigma}_g} * \Mat{\sigma}_{adv} + \Mat{\mu}_{adv}
\label{eq:f_adv}
\end{equation}
Then, the adversarial and global feature maps will together be passed to the remainder of the backbone and the FSL classifier to accomplish the $N$-way $K$-shot FSL, resulting in two predictions $\Vec{p}_g^{fsl} \in \mathbb{R}^{B \times N_c}$ and $\Vec{p}_{adv}^{fsl} \in \mathbb{R}^{NM \times N}$. Furthermore, we can get $\mathcal{L}_{fsl}$:
 \begin{equation}
     \mathcal{L}_{fsl} = \mathcal{L}_{CE}(\Vec{p}_g^{fsl}, y_{fsl}) + \mathcal{L}_{CE}(\Vec{p}_{adv}^{fsl}, y_{fsl})
\label{eq:l_ws}
 \end{equation}
 where $y_{fsl}\in \mathbb{R}^{NM}$ is the query samples' logical labels.
 \subsubsection{DCO.}
We design a novel objective function named Discrepancy \& Consistency Optimization (DCO) to maximize seen-unseen domain visual discrepancy and global-crop consistency for overall features $ \mathbb{F}_{all} = \{\Mat{F}_{1}, \Mat{F}_{2}, \cdot\cdot\cdot, \Mat{F}_{k}, \Mat{F}_g, \Mat{F}_{adv}\}$. For seen-unseen domain discrepancy maximum, we consider the global and crop features to belong to the seen domain and the generated adversarial features to belong to the unseen domain. Therefore, it is possible to make the generated adversarial features located as far away from the source domain as possible. The domain discriminator contains a dropout layer and a fully connected layer. The domain discrepancy loss is:
\begin{equation}
    \mathcal{L}_{dom} = \sum\nolimits_{\Mat{F} \in \mathbb{F}_{all}}\mathcal{L}_{CE}(f_{dom}(\Mat{F};\theta_{dom}), d_F)
\label{eq:l_dom}
\end{equation}
where $d_F \in \{0, 1\}$ is the domain label with $0$ (\textit{resp.}, $1$) indicating $\Mat{F}$ is from the seen(\textit{resp.}, the unseen) domain.
Moreover, we enforce the semantic consistency between the global and crop features as:
\begin{equation}
    \mathcal{L}_{con} = \sum_{i=1}^k(\lambda\mathcal{L}_{CE}(\Vec{p}_{i}, \Vec{p}_g) + (1-\lambda) \mathcal{L}_{CE}(\Vec{p}_{i}^{fsl}, y_{fsl}))
\end{equation}
where, $\Vec{p}_{i}^{fsl} = f_{re}(\Mat{F}_{i};\theta_{re})$. We use Kullback-Leibler divergence loss $KL(\cdot)$ to maximize global-adversarial consistency as:
\begin{equation}
    \mathcal{L}_{adv} = KL(\Vec{p}_{adv}^{fsl}, \Vec{p}_g^{fsl})
\label{eq:l_con}
\end{equation}
Then the final objective loss of \textbf{SVasP} is:
\begin{equation}
    \mathcal{L} = \mathcal{L}_{cls} + \mathcal{L}_{fsl} + \mathcal{L}_{dom} + \mathcal{L}_{con} + \mathcal{L}_{adv} 
\end{equation}

More construction details and the complete adversarial style generation pseudo-code can be found in Appendix A. 

\begin{table*}[t!]
  \centering
  \setlength{\tabcolsep}{0.74mm} 
  \caption{Quantitative comparison to state-of-the-arts methods on eight target datasets based on ResNet-10, which is pretrained on \textit{mini}ImageNet. Accuracy of 5-way 1-shot/5-shot tasks with 95 confidence interval are reported. ``\textbf{FT}" with \Checkmark means finetuning is used, vice versa. ``\textbf{Aver.}'' means ``Average Accuracy'' of the eight datasets. The optimal results are marked in \textbf{bold}.}
    \begin{tabular}{@{}llccccc|cccc|l@{}}
    \toprule
     & \textbf{Method} & \textbf{FT} & \textbf{ChestX} & \textbf{ISIC} & \textbf{EuroSAT} & \textbf{CropDisease} & \textbf{CUB} & \textbf{Cars} & \textbf{Places} & \textbf{Plantae} & \textbf{Aver.} \\
    \midrule
    \multirow{12}{*}{\rotatebox{90}{\small{\textbf{1-shot}}}} & GNN & \XSolidBrush  & 22.00\small{$\pm$0.46} & 32.02\small{$\pm$0.66} & 63.69\small{$\pm$1.03} & 64.48\small{$\pm$1.08} & 45.69\small{$\pm$0.68} & 31.79\small{$\pm$0.51} & 53.10\small{$\pm$0.80} & 35.60\small{$\pm$0.56} & 43.55  \\
    & FWT & \XSolidBrush  & 22.04\small{$\pm$0.44} & 31.58\small{$\pm$0.67} & 62.36\small{$\pm$1.05} & 66.36\small{$\pm$1.04} & 47.47\small{$\pm$0.75} & 31.61\small{$\pm$0.53} & 55.77\small{$\pm$0.79} & 35.95\small{$\pm$0.58} & 44.14  \\
    & ATA & \XSolidBrush  & 22.10\small{$\pm$0.20} & 33.21\small{$\pm$0.40} & 61.35\small{$\pm$0.50} & 67.47\small{$\pm$0.50} & 45.00\small{$\pm$0.50} & 33.61\small{$\pm$0.40} & 53.57\small{$\pm$0.50} & 34.42\small{$\pm$0.40} & 43.84  \\
    & SET-RCL & \XSolidBrush  & 22.74\small{$\pm$0.20} & 33.33\small{$\pm$0.40} & 65.53\small{$\pm$0.60} & 68.43\small{$\pm$0.50} & 46.98\small{$\pm$0.50} & 32.84\small{$\pm$0.40} & 56.93\small{$\pm$0.50} & 37.43\small{$\pm$0.40} & 45.53  \\
    & StyleAdv & \XSolidBrush  & 22.64\small{$\pm$0.35} & 33.96\small{$\pm$0.57} & 70.94\small{$\pm$0.82} & 74.13\small{$\pm$0.78} & 48.49\small{$\pm$0.72} & 34.64\small{$\pm$0.57} & 58.58\small{$\pm$0.83} & 41.13\small{$\pm$0.67} & 48.06  \\
     & \cellcolor{gray!15}{\textbf{SVasP}} & \cellcolor{gray!15}{\XSolidBrush}  & \cellcolor{gray!15}{\textbf{23.23\small{$\pm$0.35}}} & \cellcolor{gray!15}{\textbf{37.63\small{$\pm$0.58}}} & \cellcolor{gray!15}{\textbf{72.30\small{$\pm$0.82}}} & \cellcolor{gray!15}{\textbf{75.87\small{$\pm$0.73}}} & \cellcolor{gray!15}{\textbf{49.49\small{$\pm$0.72}}} & \cellcolor{gray!15}{\textbf{35.27\small{$\pm$0.57}}} & \cellcolor{gray!15}{\textbf{59.07\small{$\pm$0.81}}} & \cellcolor{gray!15}{\textbf{41.22\small{$\pm$0.62}}} & \cellcolor{gray!15}{\textbf{49.26}} \\
    \cmidrule{2-12}
    & ATA & \Checkmark  & 22.15\small{$\pm$0.20} & 34.94\small{$\pm$0.40} & 68.62\small{$\pm$0.50} & 75.41\small{$\pm$0.50} & 46.23\small{$\pm$0.50} & 37.15\small{$\pm$0.40} & 54.18\small{$\pm$0.50} & 37.38\small{$\pm$0.40} & 47.01  \\
    & StyleAdv & \Checkmark  & 22.64\small{$\pm$0.35} & 35.76\small{$\pm$0.52} & \textbf{72.92\small{$\pm$0.75}} & \textbf{80.69\small{$\pm$0.28}} & 48.49\small{$\pm$0.72} & 35.09\small{$\pm$0.55} & 58.58\small{$\pm$0.83} & 41.13\small{$\pm$0.67} & 49.41  \\
    & \cellcolor{gray!15}{\textbf{SVasP}} & \cellcolor{gray!15}{\Checkmark}  & \cellcolor{gray!15}{\textbf{23.23\small{$\pm$0.35}}}   & \cellcolor{gray!15}{\textbf{37.63\small{$\pm$0.63}}}   & \cellcolor{gray!15}{72.30\small{$\pm$0.83}}   & \cellcolor{gray!15}{77.45\small{$\pm$0.68}}   & \cellcolor{gray!15}{\textbf{49.49\small{$\pm$0.72}}}   & \cellcolor{gray!15}{\textbf{38.18\small{$\pm$0.61}}}   & \cellcolor{gray!15}{\textbf{59.07\small{$\pm$0.81}}}   & \cellcolor{gray!15}{\textbf{41.22\small{$\pm$0.62}}}   & \cellcolor{gray!15}{\textbf{49.82}} \\
    \bottomrule
    \toprule
     & \textbf{Method} & \small{\textbf{FT}} & \textbf{ChestX} & \textbf{ISIC} & \textbf{EuroSAT} & \textbf{CropDisease} & \textbf{CUB} & \textbf{Cars} & \textbf{Places} & \textbf{Plantae} & \textbf{Aver.} \\
    \midrule
    \multirow{17}{*}{\rotatebox{90}{\small{\textbf{5-shot}}}} & GNN & \XSolidBrush  & 25.27\small{$\pm$0.46} & 43.94\small{$\pm$0.67} & 83.64\small{$\pm$0.77} & 87.96\small{$\pm$0.67} & 62.25\small{$\pm$0.65} & 44.28\small{$\pm$0.63} & 70.84\small{$\pm$0.65} & 52.53\small{$\pm$0.59} & 58.84  \\
    & FWT & \XSolidBrush  & 25.18\small{$\pm$0.45} & 43.17\small{$\pm$0.70} & 83.01\small{$\pm$0.79} & 87.11\small{$\pm$0.67} & 66.98\small{$\pm$0.68} & 44.90\small{$\pm$0.64} & 73.94\small{$\pm$0.67} & 53.85\small{$\pm$0.62} & 59.77  \\
    & ATA & \XSolidBrush  & 24.32\small{$\pm$0.40} & 44.91\small{$\pm$0.40} & 83.75\small{$\pm$0.40} & 90.59\small{$\pm$0.30} & 66.22\small{$\pm$0.50} & 49.14\small{$\pm$0.40} & 75.48\small{$\pm$0.40} & 52.69\small{$\pm$0.40} & 60.89  \\
    & SET-RCL & \XSolidBrush  & 25.65\small{$\pm$0.20} & 44.93\small{$\pm$0.40} & 83.84\small{$\pm$0.40} & 88.11\small{$\pm$0.30} & 68.05\small{$\pm$0.50} & 47.95\small{$\pm$0.40} & 76.23\small{$\pm$0.40} & 54.70\small{$\pm$0.40} & 61.18  \\
    & StyleAdv & \XSolidBrush  & 26.07\small{$\pm$0.37} & 45.77\small{$\pm$0.51} & 86.58\small{$\pm$0.54} & 93.65\small{$\pm$0.39} & 68.72\small{$\pm$0.67} & 50.13\small{$\pm$0.68} & 77.73\small{$\pm$0.62} & \textbf{61.52\small{$\pm$0.68}} & 63.77  \\
    & \cellcolor{gray!15}{\textbf{SVasP}} & \cellcolor{gray!15}{\XSolidBrush}  & \cellcolor{gray!15}{\textbf{26.87\small{$\pm$0.38}}} & \cellcolor{gray!15}{\textbf{51.10\small{$\pm$0.58}}} & \cellcolor{gray!15}{\textbf{88.72\small{$\pm$0.52}}} & \cellcolor{gray!15}{\textbf{94.52\small{$\pm$0.33}}} & \cellcolor{gray!15}{\textbf{68.95\small{$\pm$0.66}}} & \cellcolor{gray!15}{\textbf{52.13\small{$\pm$0.66}}} & \cellcolor{gray!15}{\textbf{77.78\small{$\pm$0.62}}} & \cellcolor{gray!15}{60.63\small{$\pm$0.64}}& \cellcolor{gray!15}{\textbf{65.09}} \\
    \cmidrule{2-12}
    & Fine-tune & \Checkmark  & 25.97\small{$\pm$0.41} & 48.11\small{$\pm$0.64} & 79.08\small{$\pm$0.61} & 89.25\small{$\pm$0.51} & 64.14\small{$\pm$0.77} & 52.08\small{$\pm$0.74} & 70.06\small{$\pm$0.74} & 59.27\small{$\pm$0.70} & 61.00  \\
    & BSR & \Checkmark  & 26.84\small{$\pm$0.44} & 54.42\small{$\pm$0.66} & 80.89\small{$\pm$0.61} & 92.17\small{$\pm$0.45} & 69.38\small{$\pm$0.76} & 57.49\small{$\pm$0.72} & 71.09\small{$\pm$0.68} & 61.07\small{$\pm$0.76} & 64.17  \\
    & ATA & \Checkmark  & 25.08\small{$\pm$0.20} & 49.79\small{$\pm$0.40} & 89.64\small{$\pm$0.30} & 95.44\small{$\pm$0.20} & 69.83\small{$\pm$0.50} & 54.28\small{$\pm$0.50} & 76.64\small{$\pm$0.40} & 58.08\small{$\pm$0.40} & 64.85  \\
    & NSAE & \Checkmark  & 27.10\small{$\pm$0.44} & 54.05\small{$\pm$0.63} & 83.96\small{$\pm$0.57} & 93.14\small{$\pm$0.47} & 68.51\small{$\pm$0.76} & 54.91\small{$\pm$0.74} & 71.02\small{$\pm$0.72} & 59.55\small{$\pm$0.74} & 64.03  \\
    & RDC & \Checkmark  & 25.48\small{$\pm$0.20} & 49.06\small{$\pm$0.30} & 84.67\small{$\pm$0.30} & 93.55\small{$\pm$0.30} & 67.77\small{$\pm$0.40} & 53.75\small{$\pm$0.50} & 74.65\small{$\pm$0.40} & 60.63\small{$\pm$0.40} & 63.70  \\
    & StyleAdv & \Checkmark  & 26.24\small{$\pm$0.35} & 53.05\small{$\pm$0.54} & 91.64\small{$\pm$0.43} & 96.51\small{$\pm$0.28} & 70.90\small{$\pm$0.63} & 56.44\small{$\pm$0.68} & \textbf{79.35\small{$\pm$0.61}} & 64.10\small{$\pm$0.64} & 67.28  \\
    & \cellcolor{gray!15}{\textbf{SVasP}} & \cellcolor{gray!15}{\Checkmark}  & \cellcolor{gray!15}{\textbf{27.25\small{$\pm$0.39}}} & \cellcolor{gray!15}{\textbf{55.43\small{$\pm$0.59}}} & \cellcolor{gray!15}{\textbf{91.77\small{$\pm$0.41}}} & \cellcolor{gray!15}{\textbf{96.79\small{$\pm$0.26}}} & \cellcolor{gray!15}{\textbf{72.06\small{$\pm$0.65}}} & \cellcolor{gray!15}{\textbf{59.99\small{$\pm$0.69}}} & \cellcolor{gray!15}{78.91\small{$\pm$0.65}} & \cellcolor{gray!15}{\textbf{64.21\small{$\pm$0.66}}} & \cellcolor{gray!15}{\textbf{68.30}} \\
    \bottomrule
    \end{tabular}%
  \label{tab:main-results-rn10}%
\end{table*}%
\section{Experiments}
\subsection{Datasets} 
Following the BSCD-FSL benchmark proposed in BSCD-FSL \cite{guo2020broader} and the \textit{mini}-CUB benchmark proposed in FWT \cite{tseng2020cross}, we use \textit{mini}ImageNet \cite{ravi2016optimization} with 64 classes as the source domain. The target domains include eight datasets: ChestX \cite{wang2017chestx}, ISIC \cite{tschandl2018ham10000}, EuroSAT \cite{helber2019eurosat}, CropDisease \cite{mohanty2016using}, CUB \cite{wah2011caltech}, Cars \cite{krause20133d}, Places \cite{zhou2017places}, and Plantae \cite{van2018inaturalist}. In our Single Source CD-FSL setting, target domain datasets are not available during meta-training stage.
\subsection{Implementation Details}
Using ResNet-10~\cite{he2016deep} as the backbone and GNN as the $N$-way $K$-shot classifier, the network is meta-trained for 200 epochs with 120 episodes per epoch. ResNet-10 is pretrained \textit{mini}ImageNet using traditional batch training. The optimizer is Adam with a learning rate of 0.001. Additionally, using ViT-small~\cite{dosovitskiy2020image} as the feature extractor and ProtoNet~\cite{laenen2021episodes} as the $N$-way $K$-shot classifier, the network is meta-trained for 20 epochs with 2000 episodes per epoch. The optimizer is SGD with a learning rate of 5e-5 and 0.001 for $E$ and $f_{re}$, respectively. ViT-small is pretrained on ImageNet1K by DINO~\cite{caron2021emerging}. We evaluate the proposed framework during testing by average classification accuracy over 1000 episodes with a 95\% confidence interval. Each class contains 5 support samples and 15 query samples. Hyper-parameters are set as follows: $\xi = 0.1$, $k = 2$, $\lambda = 0.2$ and choose $\kappa_1$, $\kappa_2$ from $[0.008, 0.08, 0.8]$. The probability to perform style change is set to 0.2. 
The details of the finetuning are attached in Appendix A. All the experiments are conducted on a single NVIDIA GeForce RTX 3090.
\subsection{Experimental Results}
\begin{table*}[t!]
  \centering
  \setlength{\tabcolsep}{0.74mm} 
  \caption{Quantitative comparison to state-of-the-arts methods on eight target datasets based on ViT-small, which is pretrained on ImageNet1K by DINO. Accuracy of 5-way 1-shot/5-shot tasks with 95 confidence interval are reported.}
    \begin{tabular}{@{}llccccc|cccc|l@{}}
    \toprule
     & \textbf{Method} & \textbf{FT} & \textbf{ChestX} & \textbf{ISIC} & \textbf{EuroSAT} & \textbf{CropDisease} & \textbf{CUB} & \textbf{Cars} & \textbf{Places} & \textbf{Plantae} & \textbf{Aver.} \\
    \midrule
    \multirow{6}{*}{\rotatebox{90}{\small{\textbf{1-shot}}}} & StyleAdv & \XSolidBrush  & \textbf{22.92\small{$\pm$0.32}} & 33.05\small{$\pm$0.44} & 72.15\small{$\pm$0.65} & \textbf{81.22\small{$\pm$0.61}} & 84.01\small{$\pm$0.58} & 40.48\small{$\pm$0.57} & 72.64\small{$\pm$0.67} & 55.52\small{$\pm$0.66} & 57.75  \\
    & \cellcolor{gray!15}{\textbf{SVasP}} & \cellcolor{gray!15}{\XSolidBrush}  & \cellcolor{gray!15}{22.68\small{$\pm$0.30}} & \cellcolor{gray!15}{\textbf{34.49\small{$\pm$0.46}}} & \cellcolor{gray!15}{\textbf{72.50\small{$\pm$0.62}}} & \cellcolor{gray!15}{80.82\small{$\pm$0.62}} & \cellcolor{gray!15}{\textbf{85.56\small{$\pm$0.57}}} & \cellcolor{gray!15}{\textbf{40.51\small{$\pm$0.59}}} & \cellcolor{gray!15}{\textbf{75.93\small{$\pm$0.66}}} & \cellcolor{gray!15}{\textbf{56.25\small{$\pm$0.65}}} & \cellcolor{gray!15}{\textbf{58.59}} \\
    \cmidrule{2-12}
    & PMF & \Checkmark  & 21.73\small{$\pm$0.30} & 30.36\small{$\pm$0.36} & 
    70.74\small{$\pm$0.63} & 80.79\small{$\pm$0.62} & 78.13\small{$\pm$0.66} & 37.24\small{$\pm$0.57} & 71.11\small{$\pm$0.71} & 53.60\small{$\pm$0.66} & 55.46\\
    & StyleAdv & \Checkmark  & \textbf{22.92\small{$\pm$0.32}} & 33.99\small{$\pm$0.46} & 74.93\small{$\pm$0.58} & \textbf{84.11\small{$\pm$0.57}} & 84.01\small{$\pm$0.58} & 40.48\small{$\pm$0.57} & 72.64\small{$\pm$0.67} & 55.52\small{$\pm$0.66} & 58.57  \\
    & \cellcolor{gray!15}{\textbf{SVasP}} & \cellcolor{gray!15}{\Checkmark}  & \cellcolor{gray!15}{22.68\small{$\pm$0.30}} & \cellcolor{gray!15}{\textbf{34.49\small{$\pm$0.46}}} & \cellcolor{gray!15}{{\textbf{75.51\small{$\pm$0.57}}}} & \cellcolor{gray!15}{83.98\small{$\pm$0.55}} & \cellcolor{gray!15}{\textbf{85.56\small{$\pm$0.57}}} & \cellcolor{gray!15}{\textbf{40.51\small{$\pm$0.59}}} & \cellcolor{gray!15}{\textbf{75.93\small{$\pm$0.66}}} & \cellcolor{gray!15}{\textbf{56.25\small{$\pm$0.65}}} & \cellcolor{gray!15}{\textbf{59.36}} \\
    \bottomrule
    \toprule
     & \textbf{Method} & \small{\textbf{FT}} & \textbf{ChestX} & \textbf{ISIC} & \textbf{EuroSAT} & \textbf{CropDisease} & \textbf{CUB} & \textbf{Cars} & \textbf{Places} & \textbf{Plantae} & \textbf{Aver.} \\
    \midrule
    \multirow{6}{*}{\rotatebox{90}{\small{\textbf{5-shot}}}} & StyleAdv & \XSolidBrush  & \textbf{26.97\small{$\pm$0.33}} & 47.73\small{$\pm$0.44} & 88.57\small{$\pm$0.34} & \textbf{94.85\small{$\pm$0.31}} & 95.82\small{$\pm$0.27} & 61.73\small{$\pm$0.62} & 88.33\small{$\pm$0.40} & 75.55\small{$\pm$0.54} & 72.44  \\
    & \cellcolor{gray!15}{\textbf{SVasP}} & \cellcolor{gray!15}{\XSolidBrush}  & \cellcolor{gray!15}{26.77\small{$\pm$0.34}} & \cellcolor{gray!15}{\textbf{49.75\small{$\pm$0.46}}} & \cellcolor{gray!15}{\textbf{88.69\small{$\pm$0.35}}} & \cellcolor{gray!15}{93.25\small{$\pm$0.36}} & \cellcolor{gray!15}{\textbf{95.95\small{$\pm$0.23}}} & \cellcolor{gray!15}{\textbf{62.60\small{$\pm$0.61}}} & \cellcolor{gray!15}{\textbf{89.19\small{$\pm$0.39}}} & \cellcolor{gray!15}{\textbf{76.49\small{$\pm$0.50}}} & \cellcolor{gray!15}{\textbf{72.84}} \\
    \cmidrule{2-12}
    & PMF & \Checkmark  & 27.27 & 50.12 & 
    85.98 & 92.96 & - & - & - & - & -\\
    & StyleAdv & \Checkmark  & \textbf{26.97\small{$\pm$0.33}} & 51.23\small{$\pm$0.51} & 90.12\small{$\pm$0.33} & 95.99\small{$\pm$0.27} & 95.82\small{$\pm$0.27} & 66.02\small{$\pm$0.64} & 88.33\small{$\pm$0.40} & 78.01\small{$\pm$0.54} & 74.06  \\
    & \cellcolor{gray!15}{\textbf{SVasP}} & \cellcolor{gray!15}{\Checkmark}  & \cellcolor{gray!15}{26.77\small{$\pm$0.34}} & \cellcolor{gray!15}{\textbf{51.62\small{$\pm$0.50}}} & \cellcolor{gray!15}{\textbf{90.55\small{$\pm$0.34}}} & \cellcolor{gray!15}{{\textbf{96.17\small{$\pm$0.30}}}} & \cellcolor{gray!15}{\textbf{95.95\small{$\pm$0.23}}} & \cellcolor{gray!15}{\textbf{66.47\small{$\pm$0.62}}} & \cellcolor{gray!15}{\textbf{89.19\small{$\pm$0.39}}} & \cellcolor{gray!15}{\textbf{78.67\small{$\pm$0.52}}} & \cellcolor{gray!15}{\textbf{74.42}} \\
    \bottomrule
    \end{tabular}%
  \label{tab:main-results-vit}%
\end{table*}%
\subsubsection{Comparison to SOTA methods on ResNet-10.}We compare the proposed SVasP with state of the art methods with ResNet-10 as the backbone in Table \ref{tab:main-results-rn10}. For a fair comparison, all the competing methods follow the single source training scheme, which is more realistic and difficult. Concretly, nine representative single source CD-FSL methods are introduced including GNN \cite{garcia2018few}, FWT \cite{tseng2020cross}, ATA \cite{wang2021cross}, SET-RCL \cite{zhang2022free}, StyleAdv \cite{fu2023styleadv}, Fine-tune \cite{guo2020broader}, BSR \cite{liu2020feature}, NSAE \cite{liang2021boosting} and RDC \cite{li2022ranking}. 
As shown, under whether setting, our method outperforms the second-best approach in terms of average accuracy with a clear margin and builds a new state of the art in the majority of domains. More precisely, under 1-shot setting on ResNet-10, SVasP performs better in all domains and surpasses the strongest competitor StyleAdv significantly by $+0.59\%$, $+3.67\%$, $+1.36\%$, $+1.74\%$, $+1.00\%$, $+0.63\%$ on ChestX, ISIC, EuroSAT, CropDisease, CUB, Cars, respectively. Under 5-shot setting on ResNet-10, SVasP performs better in 7 out of 8 domains, and the superiority of SVasP is even larger with higher accuracy by $+0.80\%$, $+5.33\%$, $+2.14\%$, $+0.87\%$, $+2.00\%$ on ChestX, ISIC, EuroSAT, CropDisease, Cars, respectively. Despite being trained on one dataset, SVasP has good generalization ability, thus producing the optimal style-based augmentation policies for the unseen target domains.
\begin{table}[t!]
  \centering
  \setlength{\tabcolsep}{3.5mm} 
  \caption{Ablation study of the proposed method with different component combinations. ``SV'' indicates SV Gradient Ensemble module.}
    \begin{tabular}{c|ccc|c}
    \toprule
    Method & \textbf{SV} & $\mathcal{L}_{dom}$ & $\mathcal{L}_{con}$ & \textbf{Aver. (\%)} \\
    \midrule
    Baseline & - & - & - & 62.07 \\
    \midrule
    \multirow{4}{*}{\small{Proposed}} & \Checkmark  &  &  & 62.61 \\
    & \Checkmark  & \Checkmark &  & 63.69 \\
    & \Checkmark  &  & \Checkmark & 64.05 \\
    & \Checkmark  & \Checkmark & \Checkmark & \textbf{65.09} \\
    \bottomrule
    \end{tabular}%
  \label{tab:ab-3module}%
\end{table}%
\subsubsection{Comparison to SOTA methods on ViT-small.} To further evaluate the effectiveness of our proposed technique, we apply the proposed SVasP idea to ViT models and compare their performance over other methods on the eight datasets with ViT-small as the backbone and ProtoNet as the classifier. As shown in Table \ref{tab:main-results-vit}, our SVasP is compared with methods like StyleAdv and PMF. SVasP achieves 58.59\% and 72.84\% top 1 average accuracy on either 5-way 1-shot or 5-way 5shot setting, which outperforms StyleAdv by 0.84\%, 0.40\%, respectively.

\subsection{Qualitative Evaluation}
We have performed an exhaustive and fair experimental analysis of the proposed method SVasP and the experimental results with ResNet-10 as the backbone and GNN as the classifier under the 5-way 5-shot setting are reported. More experimental results are attached in Appendix B.
\begin{figure}[t!]
\centering
\includegraphics[width=\linewidth]{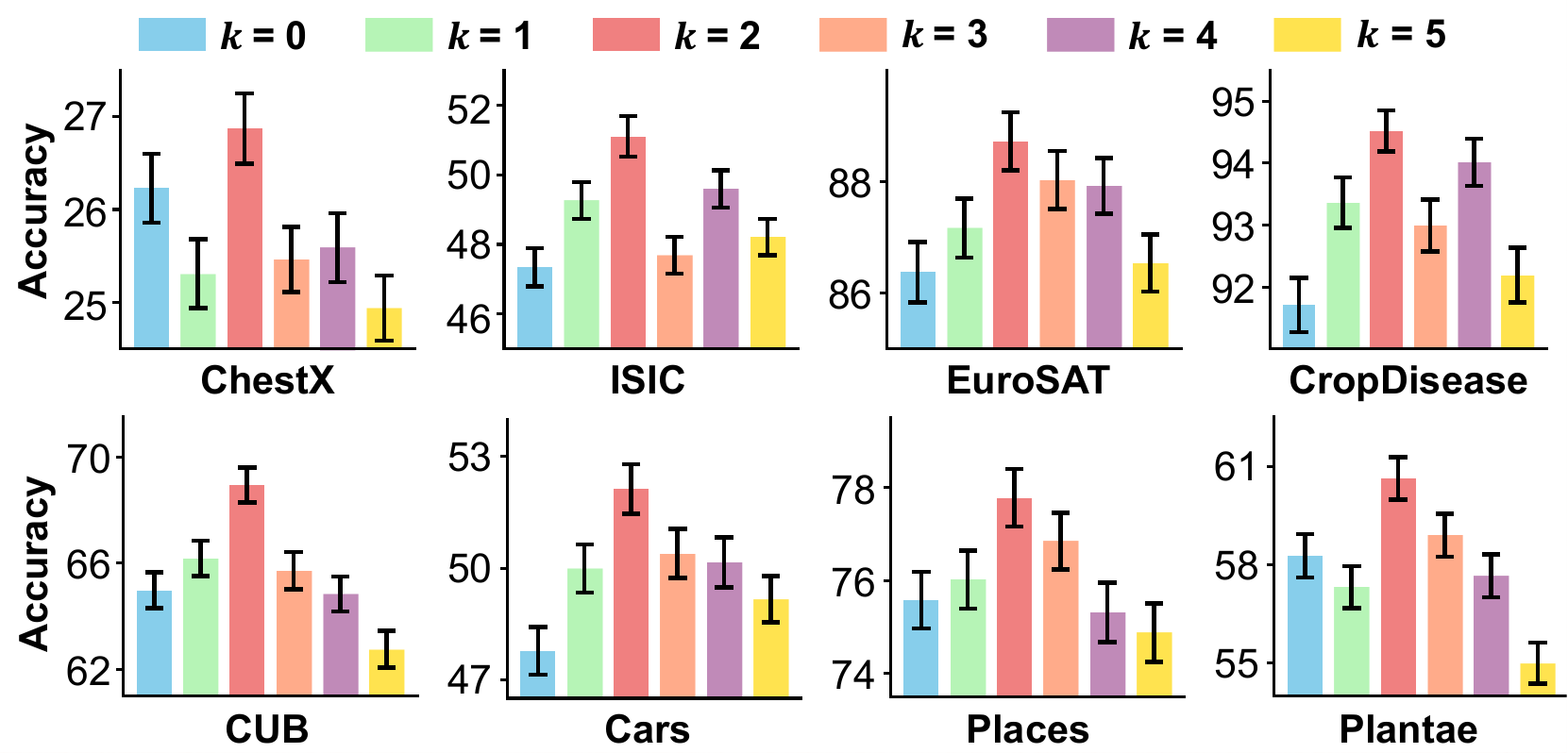} 
\caption{Performances on different numbers of crops $k$.}
\label{fig:k_crop}
\end{figure}
\subsubsection{Impact of different component in SVasP.} To investigate the contribution of different components, we perform an ablation study on SVasP and report the result of average accuracy on eight target domains in Table \ref{tab:ab-3module}. Specifically, we study the main technical contributions by (a) whether using SV (means the SV Gradient Ensemble module), (b) whether $\mathcal{L}_{dom}$ and (c) whether $\mathcal{L}_{con}$. Among these variants, we can find that the SV Gradient Ensemble module effectively utilizes the source domain style gradients to alleviate the domain shift problem. With well constrained $\mathcal{L}_{dom}$ and $\mathcal{L}_{con}$, SVasP improves the generalization performance and substantially improves the accuracy up to 3.02\% on average. 
\subsubsection{Impact of different crop numbers $k$.} We investigate the optimal solution for the number of crops and find that the model is most robust when the number is set to 2, as illustrated in Figure \ref{fig:k_crop}. Because an insufficient number of crops (\eg, 0, 1) fail to represent the style gradients of the source domain and stabilize the global style perturbation. Moreover, excessive crops (\eg, 3, 4, 5) can lead to overfitting of the model and limited by the source domain style. 

\subsubsection{Impact of different strategies for $\xi$ and $\lambda$.} 
\begin{figure}[t!]
\centering
\includegraphics[width=0.47\textwidth]{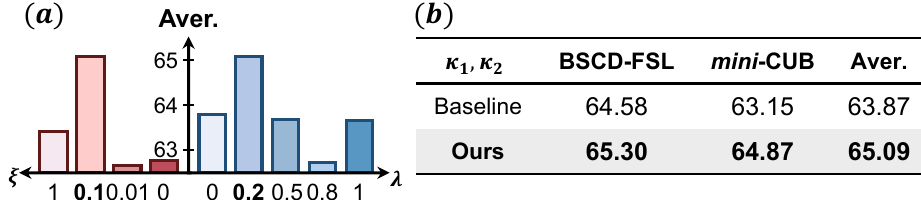} 
\caption{Performances on (a) different $\xi$, $\lambda$ and (b) whether use same $\kappa_1$, $\kappa_2$. The average accuracy (\%) is reported. }
\label{fig:hyper}
\end{figure}

The decay factor $\xi$ controls the proportion of the crop style gradients that are incorporated into the global style gradients. In addition, the main component crop consistency loss $\mathcal{L}_{con}$ has a large impact on the performance of the model, which consists of the global-crop prediction consistency loss and the crop FSL classification loss. Performances on different $\lambda$ and $\xi$ are illustrated in Figure \ref{fig:hyper} (a).
As shown, the accuracy rises as $\xi$ increase from $0$ to $0.1$, as the proportion increases and provides more source domain gradients. However, the accuracy decreases when the proportion is 1, as too high a proportion of the crop style gradients leads to weak global style gradients. For $\lambda$, setting the value of $\lambda$ to 0.2 can realize an increase in the mean classification accuracy compared to other settings of approximately 1.62\%.

\subsubsection{Impact of different selection methods for $\kappa_1$ and $\kappa_2$.} Unlike styleadv, which sets $\kappa_1$ and $\kappa_2$ to the same value, we allow $\kappa_1$ and $\kappa_2$ to have different values to diversify the style. The experimental results verify the rationality of our setup, as shown in Figure \ref{fig:hyper} (b).
\begin{figure}[t]
\centering
\includegraphics[width=\linewidth]{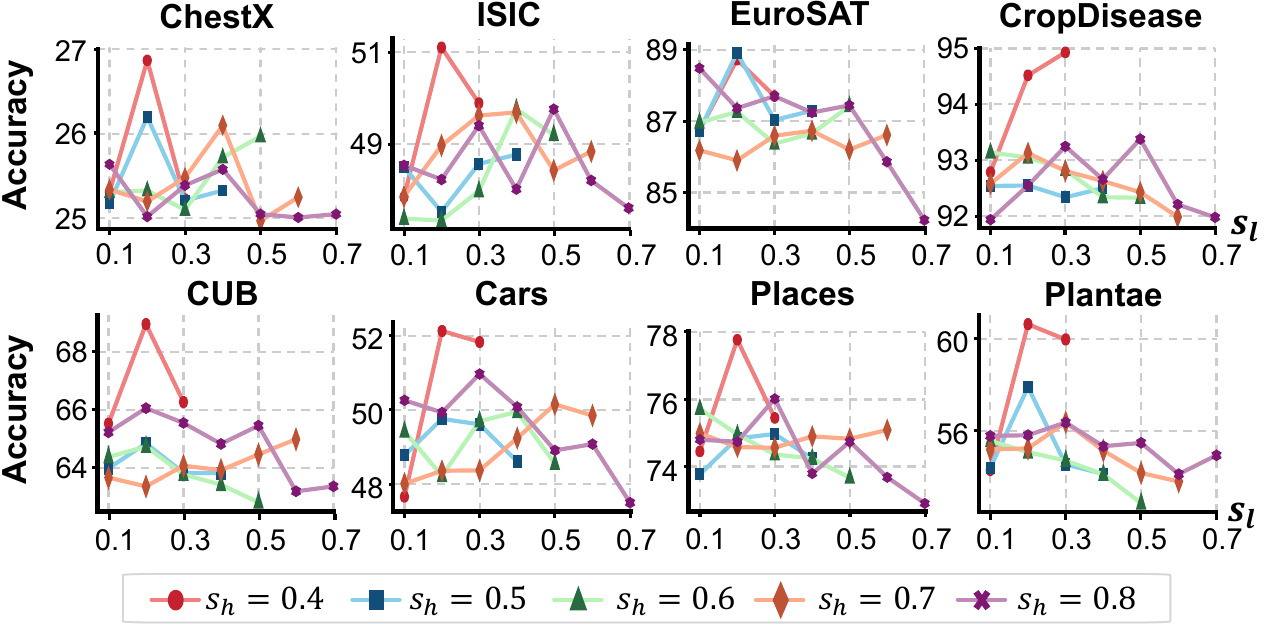} 
\caption{Performances on different scale parameters $s$.}
\label{fig:scale}
\end{figure}

\subsubsection{Impact of different scale parameters $s$.} We evaluate the impact of different scale parameters $s$, which determines the the area of the crop images. It's important to study optimal values of $s$ because when the area is large, the model overlooks the local regions of inputs. We investigate the performances with $s_l \in [0.1, 0.2, 0.3, 0.4, 0.5, 0.6, 0.7]$ and $s_h \in [0.4, 0.5, 0.6, 0.7, 0.8]$, with $s_l < s_h$. The optimal result is reached when $s = (0.2, 0.4)$, as shown in Figure \ref{fig:scale}. We observe that SVasP with smaller area of crop images performs better, which demonstrates the effectiveness of our introduction of localized crop style gradients. 

\begin{figure}[t]
\centering
\includegraphics[width=\linewidth]{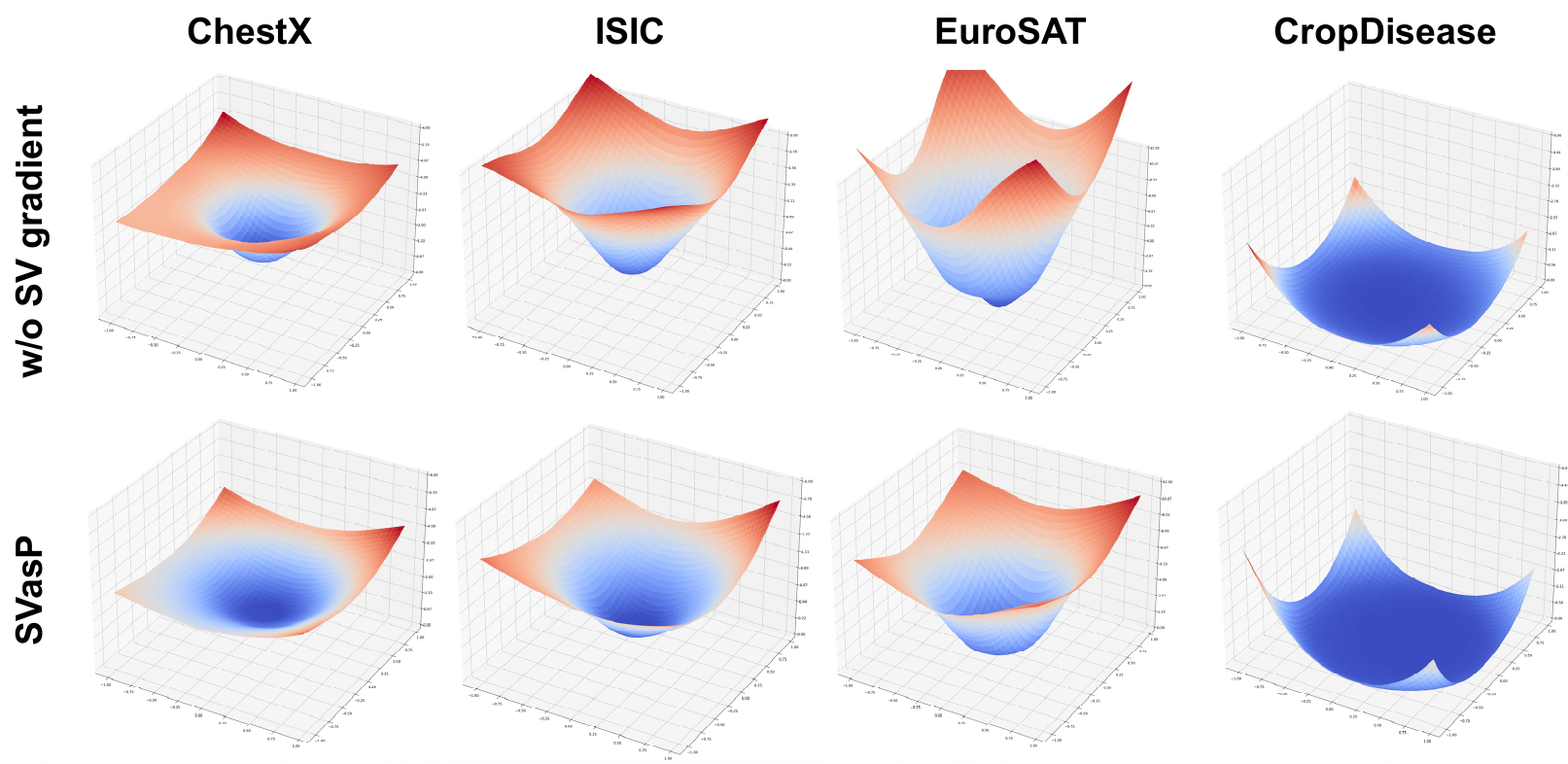} 
\caption{Loss landscape visualization results of the model without SV gradient ensemble module (first row) and our SVasP model (second row) on the  BSCD-FSL benchmark.} 
\label{fig:landscape}
\end{figure}

\begin{figure}[t]
\centering
\includegraphics[width=\linewidth]{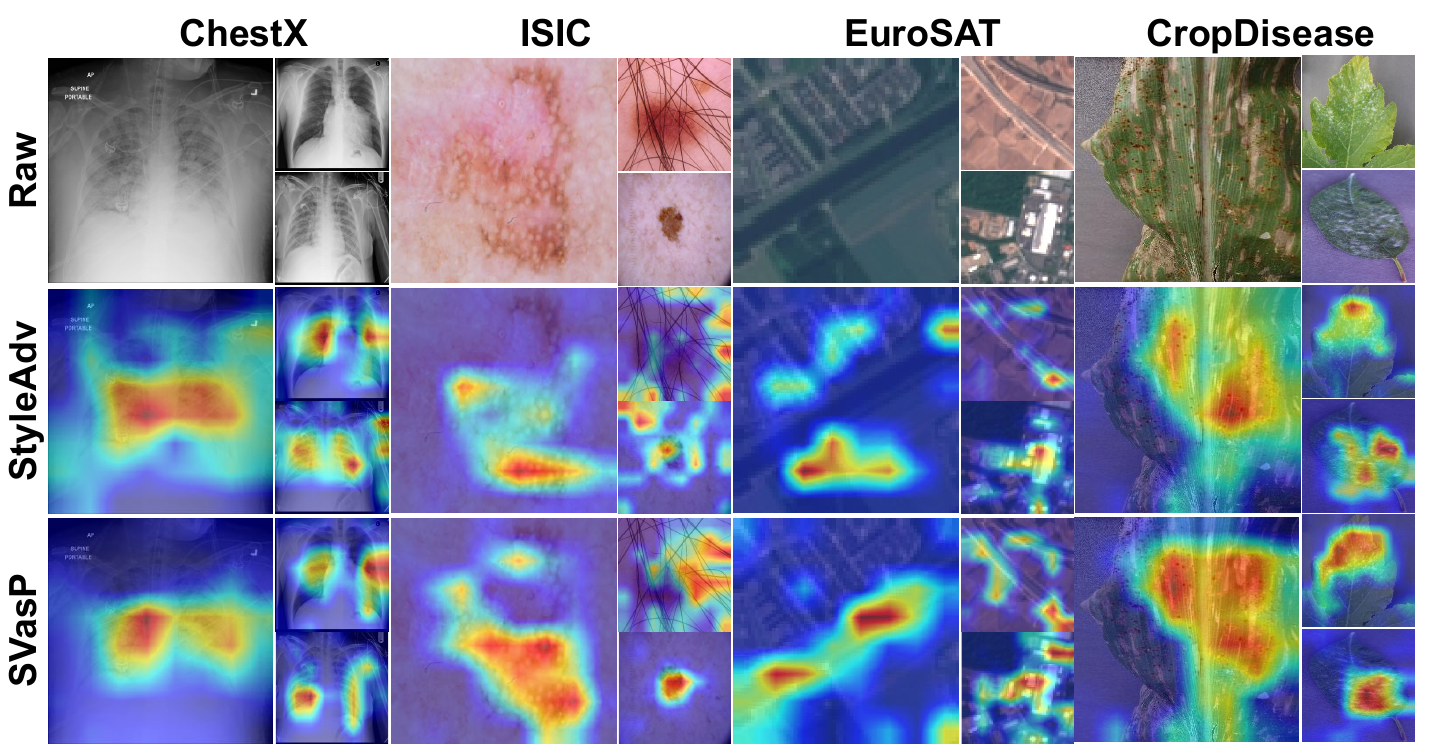} 
\caption{Grad-CAM visualization results of the StyleAdv model and our SVasP model on the BSCD-FSL benchmark. For each target dataset, three examples are demonstrated.}
\label{fig:cam}
\end{figure}

\subsubsection{Visualization Results.} We visualize the loss landscape following \cite{li2018visualizing} on the BSCD-FSL bechmark to verify the validity of our proposed important module, as shown in Figure \ref{fig:landscape}. SVasP achieves a stronger flatness which can stand for the better generalization. In additon, in order to provide a more intuitive comparison about the performance of ``SVasP''(ours) and ``StyleAdv'' model, we visualize the class-activation map using the Grad-CAM \cite{selvaraju2017grad} on the BSCD-FSL benchmark, as shown in Figure \ref{fig:cam}. We can observe that, StyleAdv may pay attention to insignificant things and is disorganized. In contrast, SVasP can focus on more key areas of the target images with the help of the localized crop style gradients. Visualization results on the \textit{mini}-CUB benchmark can be found in Appendix C.
\section{Conclusion} \label{sec:conclusion}
We explore the Single Source Cross-Domain Few-Shot Learning, focusing on the limitations of style-based approaches and addressing the domain shift problem. Our study introduces a novel network to capitalize on the localized crop style gradients, achieving state-of-the-art performance on both ResNet-10 and ViT-small backbone. To enhance the training process, we employ a random cropping strategy and integrate crop style gradients as the style perturbation stabilizers. This approach prevents the model from being confined to the source domain style and local loss minima. Extensive experimental results demonstrate the effectiveness and insights of the proposed method, highlighting its rationality and potential for broader application.
\bibliography{aaai25}

\clearpage
\appendix
\twocolumn[
\begin{center}
    {\Large \bf Supplementary Material for `` SVasP: Self-Versatility Adversarial Style Perturbation for Cross-Domain Few-Shot Learning ''} 
\end{center}
]
\noindent In the supplementary material, we provide:

\begin{itemize}
    \item More implementation details of the proposed methods.
    \item More experimental results of ablation studies.
    \item More visualization results for model evaluation.
\end{itemize}

\section{A. More Implementation Details}
\subsubsection{A.1. Model-Based Style Generation.}  \leavevmode  \\
\begin{figure}[ht]
    \centering
    \includegraphics[width=0.47\textwidth]{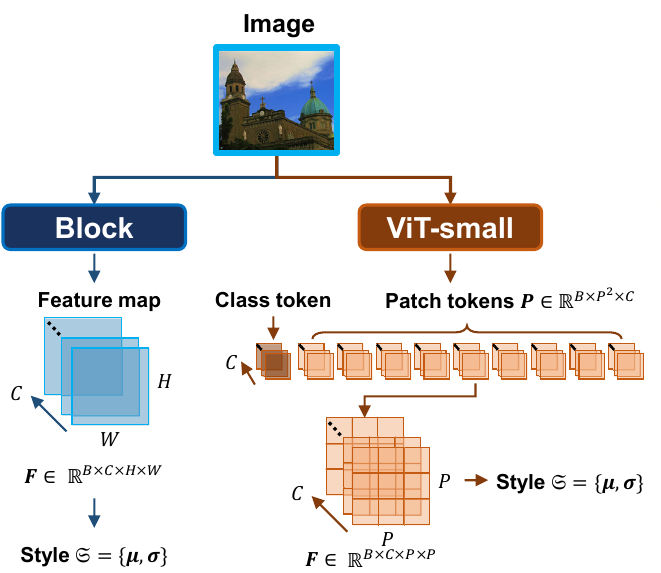}
    \caption{Comparison of style generation methods for different backbones.}
    \label{fig:supp_block-vit}
\end{figure}

The main basis of our approach is to represent the style of the domain, in addition our experiments involved two different backbones, so we detail the style generation mechanism for the different backbones. 
 
 Specifically, for the ResNet-10 backbone, styles of four blocks' features are generated for iterative adversarial style generation, separately. For each block, We get the feature map $\Mat{F} \in \mathbb{R}^{B \times C \times H \times W}$, where $B$, $C$, $H$ and $W$ denote the batch size, channel, height, width of the feature maps $\Mat{F}$. 

 For the ViT-small backbone, the input goes through the ViT-small backbone and outputs class tokens and patch tokens $\Mat{P}$, we ignore class tokens and use only patch tokens $\Mat{P}$. We then reshape the patch tokens $\Mat{P} \in \mathbb{R}^{B \times P^2 \times C}$ into the feature map $\Mat{F} \in \mathbb{R}^{B \times C \times H \times W}$, where $H = W = P$.

 At this point, we get the corresponding feature maps $\Mat{F}$ for the two backbones, as shown in Figure \ref{fig:supp_block-vit}. Thus, we can get the style $\mathfrak{S} = \{\Mat{\mu}, \Mat{\sigma}\}$ by the equations:
\begin{equation}\label{eq:supp_mu}
\Mat{\mu}=\frac{1}{HW}\sum_{h=1}^{H}\sum_{w=1}^{W}\Mat{F}_{B,C,h,w} ,
\end{equation}
\begin{equation}\label{eq:supp_sigma}
\negthickspace \negthickspace \Mat{\sigma}=\sqrt{\frac{1}{HW}\sum_{h=1}^{H}\sum_{w=1}^{W}(\Mat{F}_{B,C,h,w}-\Mat{\mu})^{2} + \epsilon},
\end{equation}
where $\epsilon$ is a small value to avoid division by zero. 
\begin{algorithm}[htbp]
\caption{SVasP attack method}
\label{alg:algorithm}
\textbf{Input}: Benign image $\Vec{x}$, target label $y$, backbone $E$ with four blocks $\{B_j\}_{j=1}^4$, standard cross-entropy loss $\mathcal{L}_{CE}(\cdot, \cdot)$. \\
\textbf{Parameter}: Scale parameter $s = \{s_l, s_h\}$, decay factor $\xi$, attack parameter $\kappa_1$ and $\kappa_2$. \\
\textbf{Output}: Adversarial style set $\mathbb{S}_{adv}=\{\mathfrak{S}_{adv}^1, \mathfrak{S}_{adv}^2, \mathfrak{S}_{adv}^3\}$. \\
\begin{algorithmic}[1] 
\STATE Random crop and resize $k$ crop images of $\Mat{x}$ and get the input set: $\mathbb{I} = \{\Mat{c}_1, \Mat{c}_2, \cdot \cdot \cdot, \Mat{c}_k, \Mat{x}\}$.
\FOR{$j = 1$ to $3$}
\FOR{$\Mat{I}$ in $\mathbb{I}$}
\STATE $\Mat{F} = \mathfrak{T}_j(\mathfrak{T}_{j-1}(\cdot\cdot\cdot(\mathfrak{T}_1({\Mat{I}, \mathfrak{S}_{adv}^1}), \cdot\cdot\cdot), \mathfrak{S}_{adv}^{j-1}), \mathfrak{S}_{adv}^j)$ \\
\STATE Calculate the style $\mathfrak{S} = \{\Mat{\mu}, \Mat{\sigma}\}$ of $\Mat{F}$ by Eq.\eqref{eq:supp_mu}, \eqref{eq:supp_sigma}
\STATE Get the prediction $\Vec{p}=f_{g}(B_4(\cdot\cdot\cdot (B_{j+1}(\Mat{F})));\theta_{g})$
\STATE $\mathcal{L}_{std} = \mathcal{L}_{std} + \mathcal{L}_{CE}(\Vec{p}, {y})$
\STATE Calculate the style gradient $\{\nabla_{\Mat{\mu}}\mathcal{L}_{std}, \nabla_{\Mat{\sigma}}\mathcal{L}_{std}\}$
\ENDFOR
\STATE Get the total style gradient set by Eq. \eqref{eq:supp_g_mu} and \eqref{eq:supp_g_sigma}.
\STATE Get the aggregated crop style gradients by Eq. \eqref{eq:supp_g_crop_mu} and \eqref{eq:supp_g_crop_sigma}. 
\STATE Get the ensemble style gradients by Eq.\eqref{eq:supp_g_e_mu} and \eqref{eq:supp_g_e_sigma}.
\STATE Get the adversarial styles of $B_j$ by Eq. \eqref{eq:supp_mu_init}, \eqref{eq:supp_std_init}, \eqref{eq:supp_mu_adv} and \eqref{eq:supp_std_adv}.
\ENDFOR
\RETURN $\Mat{\mu}_{adv}^1$, $\Mat{\sigma}_{adv}^1$, $\Mat{\mu}_{adv}^2$, $\Mat{\sigma}_{adv}^2$, $\Mat{\mu}_{adv}^3$, $\Mat{\sigma}_{adv}^3$
\end{algorithmic}
\end{algorithm}
 \begin{figure*}[htbp]
    \centering
    \includegraphics[width=1\textwidth]{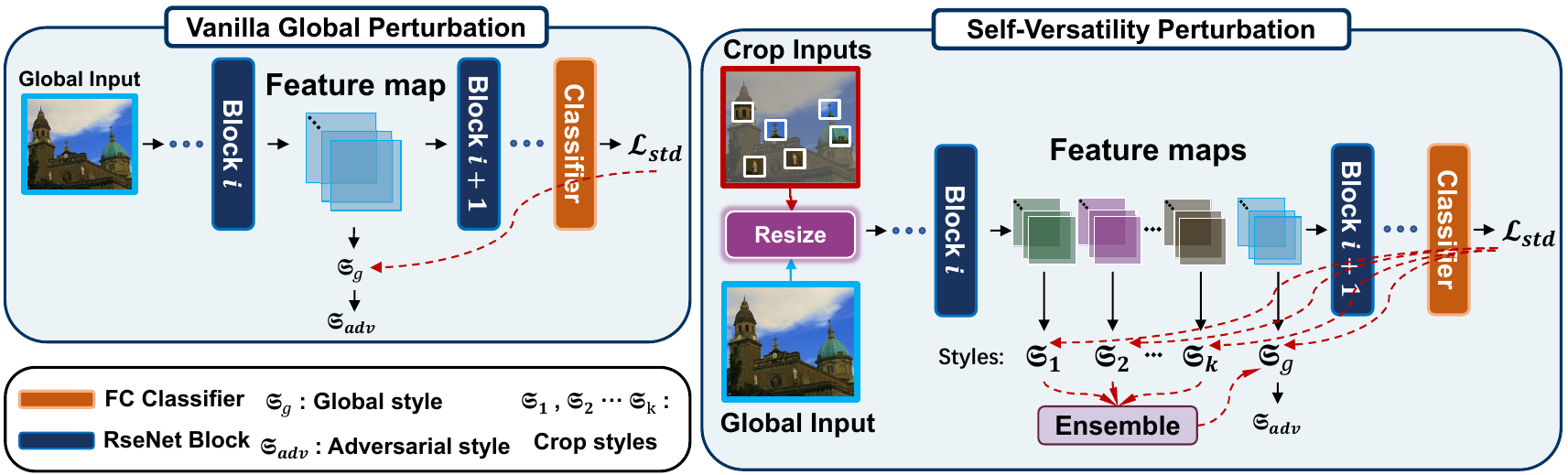}
    \caption{Comparison of the vanilla global perturbation and our self-versatility perturbation methods.}
    \label{fig:supp_crop-attack}
\end{figure*}
\subsubsection{A.2. Self-Versatility Gradient Ensemble Perturbation.}  \leavevmode  \\
 \indent One of our key contributions is the ensemble of the style gradients of crop images with the image itself, which we called self-versatility. To better help understand our proposed method, we compare it with the vanilla global style perturbation method, as illustrated in Figure \ref{fig:supp_crop-attack}. 
 The direct vanilla global style perturbation methods produce more homogeneous styles, and the use of source domain style gradients is not maximized, limiting style diversity. In contrast, our proposed method covers more information about the style of the source domain, which makes the generated adversarial style more domain-independent and generalizable.

 Moreover, as iterative synthesizing strategy is confirmed to be effective in style attack, we present the novel SVasP adversarial training method progressively. The complete adversarial style generation pseudo-code is shown in Algorithm \ref{alg:algorithm}.
The formulas used in the pseudo-code are listed below:
\begin{equation}
\begin{aligned}
\mathbb{G}^{\mu} & = \{\boldsymbol{\mathcal{G}}^{\mu}_{1}, \boldsymbol{\mathcal{G}}^{\mu}_{2}, \cdot\cdot\cdot, \boldsymbol{\mathcal{G}}^{\mu}_{k}, \boldsymbol{\mathcal{G}}^{\mu}_{g}\} \\
        & = \{\nabla_{\Mat{\mu}_{1}}\mathcal{L}_{std}, \nabla_{\Mat{\mu}_{2}}\mathcal{L}_{std}, \cdot\cdot\cdot, \nabla_{\Mat{\mu}_{k}}\mathcal{L}_{std}, \nabla_{\Mat{\mu}_{g}}\mathcal{L}_{std}\}
\end{aligned}
\label{eq:supp_g_mu}
\end{equation}
\begin{equation}
\begin{aligned}
\mathbb{G}^{\sigma} & = \{\boldsymbol{\mathcal{G}}^{\sigma}_{1}, \boldsymbol{\mathcal{G}}^{\sigma}_{2}, \cdot\cdot\cdot, \boldsymbol{\mathcal{G}}^{\sigma}_{k}, \boldsymbol{\mathcal{G}}^{\sigma}_{g}\} \\
        & = \{\nabla_{\Mat{\sigma}_{1}}\mathcal{L}_{std}, \nabla_{\Mat{\sigma}_{2}}\mathcal{L}_{std}, \cdot\cdot\cdot, \nabla_{\Mat{\sigma}_{k}}\mathcal{L}_{std}, \nabla_{\Mat{\sigma}_{g}}\mathcal{L}_{std}\}
\end{aligned}
\label{eq:supp_g_sigma}
\end{equation}
\begin{equation}
    \boldsymbol{\mathcal{G}}_{c}^\mu = Norm(\frac{1}{k} \sum(\boldsymbol{\mathcal{G}}^{\mu}_{1} + \boldsymbol{\mathcal{G}}^{\mu}_{2} + \cdot\cdot\cdot + \boldsymbol{\mathcal{G}}^{\mu}_{k}))
\label{eq:supp_g_crop_mu}
\end{equation}
\begin{equation}
    \boldsymbol{\mathcal{G}}_{c}^\sigma = Norm(\frac{1}{k} \sum(\boldsymbol{\mathcal{G}}^{\sigma}_{1} + \boldsymbol{\mathcal{G}}^{\sigma}_{2} + \cdot\cdot\cdot + \boldsymbol{\mathcal{G}}^{\sigma}_{k}))
\label{eq:supp_g_crop_sigma}
\end{equation}
\begin{equation}
    \boldsymbol{\mathcal{G}}_{e}^\mu = Norm(\boldsymbol{\mathcal{G}}_g^{\mu}) + \xi \odot \boldsymbol{\mathcal{G}}_{c}^\mu
\label{eq:supp_g_e_mu}
\end{equation}
\begin{equation}
    \boldsymbol{\mathcal{G}}_{e}^\sigma = Norm(\boldsymbol{\mathcal{G}}_g^{\sigma}) + \xi \odot \boldsymbol{\mathcal{G}}_{c}^\sigma
\label{eq:supp_g_e_sigma}
\end{equation}
\begin{equation}
    \Mat{\mu}_{init} = \Mat{\mu}_g + \varepsilon \cdot \mathcal{N}(0,I)
\label{eq:supp_mu_init}
\end{equation}
\begin{equation}
    \Mat{\sigma}_{init} = \Mat{\sigma}_g + \varepsilon \cdot \mathcal{N}(0,I)
\label{eq:supp_std_init}
\end{equation}
\begin{equation}
    \Mat{\mu}_{adv} = \Mat{\mu}_{init} + \kappa_1 \cdot sign(\boldsymbol{\mathcal{G}}_{e}^\mu)
\label{eq:supp_mu_adv}
\end{equation}
\begin{equation}
    \Mat{\sigma}_{adv} = \Mat{\sigma}_{init} + \kappa_2 \cdot sign(\boldsymbol{\mathcal{G}}_{e}^\sigma)
\label{eq:supp_std_adv}
\end{equation}
 \subsubsection{A.3. More Details for DCO.}  \leavevmode  \\
 Given the clean inputs and the adversarial features after perturbation, DCO consists of three sub losses: the domain discrepancy loss $\mathcal{L}_{dom}$, the global-crop consistency loss $\mathcal{L}_{con}$ and the global-adversarial consistency loss $\mathcal{L}_{adv}$. \\
 \textbf{Domain Discrepancy Loss.} The $\mathcal{L}_{dom}$ is introduced to distinguish between seen and unseen domains to keep the adversarial style as far away from the source domain restrictions as possible. For global and crop features, they are all classified to the seen domain and thus their domain labels $d_F$ are all 0. Besides, our main goal is to make the generated adversarial features more migratory and more unseen domain in nature, so we classify them to the unseen domain with domain labels $d_F = 1$. Thus, the domain discriminator in the module is a binary header, consisting of one fully connected layer. For each final feature passed through the backbone $\Vec{F} \in \mathbb{R}^{B \times C}$, we can get the domain prediction $\Vec{p}_{dom} = f_{dom}(\Vec{F};\theta_{dom})$, where $\Vec{p}_{dom} \in \mathbb{R}^{B \times 2}$. Thus, we can get:
 \begin{equation}
    \mathcal{L}_{dom} = \mathcal{L}_{CE}(\Vec{p}_{dom}, d_F)
\label{eq:supp_l_dom}
\end{equation}
\textbf{Crop Consistency Loss.} The $\mathcal{L}_{con}$ is introduced to restrict crop inputs. On one hand, crop and global prediction need to be semantically unified. Specifically, for the $k$ crop images, the global-crop consistency loss can be calculated by the following equation:
 \begin{equation}
    \mathcal{L}_{cg} =  \sum_{i=1}^k\mathcal{L}_{CE}(\Vec{p}_{i}, \Vec{p}_{g})
\label{eq:supp_l_cg}
\end{equation}
On the other hand, crop inputs need to achieve $N$-way $K$-shot Few-shot classification. Instead of using the benign labels, meta-laerning adopts $N$-way $K$-shot logical labels. Specifically, for the image which belongs to the $i$ class of the $N$ classes, the $N$-way $K$-shot logical label of the image is set as $y_{ws}$. Since we random crop the global images while preserve the semantics, the crop images still belong to the same logic label. Thus, the $N$-way $K$-shot loss for crop inputs is defined as:
 \begin{equation}
    \mathcal{L}_{c}^{ws} =  \sum_{i=1}^k\mathcal{L}_{CE}(\Vec{p}_{i}, y_{ws})
\label{eq:supp_l_cws}
\end{equation}
Then, we can get the final crop consistency loss with the hyper-parameter $\lambda$:
\begin{equation}
\begin{aligned}
    & \mathcal{L}_{con} = \lambda \mathcal{L}_{cg} + (1-\lambda) \mathcal{L}_{c}^{ws} \\
    & = \sum_{i=1}^k(\lambda\mathcal{L}_{CE}(\Vec{p}_{i}, \Vec{p}_g) + (1-\lambda) \mathcal{L}_{CE}(\Vec{p}_{i}^{ws}, y_{ws}))
\end{aligned}
\end{equation}
\textbf{Global-Adversarial Consistency Loss.} The $\mathcal{L}_{adv}$ is introduced to constrain the prediction $\Vec{p}_{adv}^{ws}$ and $\Vec{p}_{g}^{ws}$ by the Kullback-Leibler divergence loss, which is calculated by:
\begin{equation}
    \mathcal{L}_{adv} = \frac{1}{NM*N} \sum_{i=1}^{NM}\sum_{j=1}^N {\Vec{p}^{ws}_{g_{ij}}} * log \frac{\Vec{p}^{ws}_{g_{ij}}}{\Vec{p}^{ws}_{adv_{ij}}}
\end{equation}
\subsubsection{A.4. Details for Benchmarks and Datasets.}  \leavevmode  \\
The BSCD-FSL benchmark is proposed in BSCD-FSL \cite{guo2020broader} and the \textit{mini}-CUB benchmark is proposed in FWT \cite{tseng2020cross}.
BSCD-FSL \cite{guo2020broader} benchmark: Broader Study of Cross-Domain Few-Shot Learning (BSCD-FSL) benchmark includes image data from a diverse assortment of image acquisition methods. There are five datasets which are \textit{mini}ImageNet \cite{ravi2016optimization}, ChestX \cite{wang2017chestx}, ISIC \cite{tschandl2018ham10000}, EuroSAT \cite{helber2019eurosat} and CropDiseases \cite{mohanty2016using}.
\begin{itemize}
\item \textit{mini}ImageNet \cite{ravi2016optimization}: A dataset consists of 60000 images in total, evenly distributed across 100 classes.
\item ChestX \cite{wang2017chestx}: A medical imaging dataset which comprises 108,948 frontal-view X-ray images of 32,717 unique patients with the text-mined eight disease image labels.
\item ISIC \cite{tschandl2018ham10000}: A dataset published by the International Skin Imaging Collaboration as a large-scale dataset of dermoscopy images.
\item EuroSAT \cite{helber2019eurosat}: A dataset based on Sentinel-2 satellite images covering 13 spectral bands and consisting out of 10 classes.
\item CropDiseases \cite{mohanty2016using}: A dataset consists of about 87K RGB images of healthy and diseased crop leaves which is categorized into 38 different classes.
\end{itemize}
\textit{mini}-CUB \cite{tseng2020cross}: min-CUB benchmark is proposed in FWT \cite{tseng2020cross}, including five datasets which are \textit{mini}ImageNet \cite{ravi2016optimization}, CUB \cite{wah2011caltech}, Cars \cite{krause20133d}, Places \cite{zhou2017places} and Plantae \cite{van2018inaturalist}. In this benchmark, \textit{mini}ImageNet is always regarded as the source domain and others datasets are regarded as the target domains. 
\begin{itemize}
\item CUB \cite{wah2011caltech}: A dataset contains 200 different categories of bird images.
\item Cars \cite{krause20133d}: The Stanford Cars dataset consists of 196 classes of cars with a total of 16,185 images.
\item Places \cite{zhou2017places}: A dataset contains over 10 millions labeled exemplars from 434 place categories.
\item Plantae \cite{van2018inaturalist}: Plantae dataset is one of dataset iNat2017. There are 2101 categories and 196613 images in this dataset.
\end{itemize}
\subsubsection{A.5. Details for Finetuning.}  \leavevmode  \\
We follow the finetune setting in StyleAdv \cite{fu2023styleadv} to ensure that the comparison is fair. Specifically, we finetune the meta-trained model with pseudo training episodes. The specific finetuning details are shown in Table \ref{tab:finetune}.Compared with 5-way 1-shot tasks, 5-way 5-shot tasks need more training iterations because more query images are required to be classified. Compared with RseNet-10 backbone, ViT-small backbone need smaller learning rate bacause ViT-small model are pretrained on the large-scale model.
 \begin{table}[ht]
  \centering
  \setlength{\tabcolsep}{2mm} 
    \begin{tabular}{c|c|c|c|c}
    \toprule
    \textbf{Backbone} & \textbf{Task} & \textbf{Opt.} & \textbf{Iter.} & \textbf{lr} \\
    \midrule
    ResNet-10 & 5-way 1-shot & Adam & 10 & \{0, 0.005\} \\
    ResNet-10 & 5-way 5-shot & Adam & 50 & \{0, 0.001\} \\
    \midrule
    ViT-small & 5-way 1-shot & SGD & 20 & \{0, 5e-5\} \\
    ViT-small & 5-way 5-shot & SGD & 50 & \{0, 5e-5\} \\
    \bottomrule
    \end{tabular}%
  \caption{The finetuning deatails for ResNet-10 and ViT-small backbones. The ``Opt.'', the ``Iter.'' and the ``lr'' represent the optimizer, the finetuning iterations and the learning rate, respectively. }
  \label{tab:finetune}%
\end{table}%
\begin{figure}[t!]
\centering
\includegraphics[width=\linewidth]{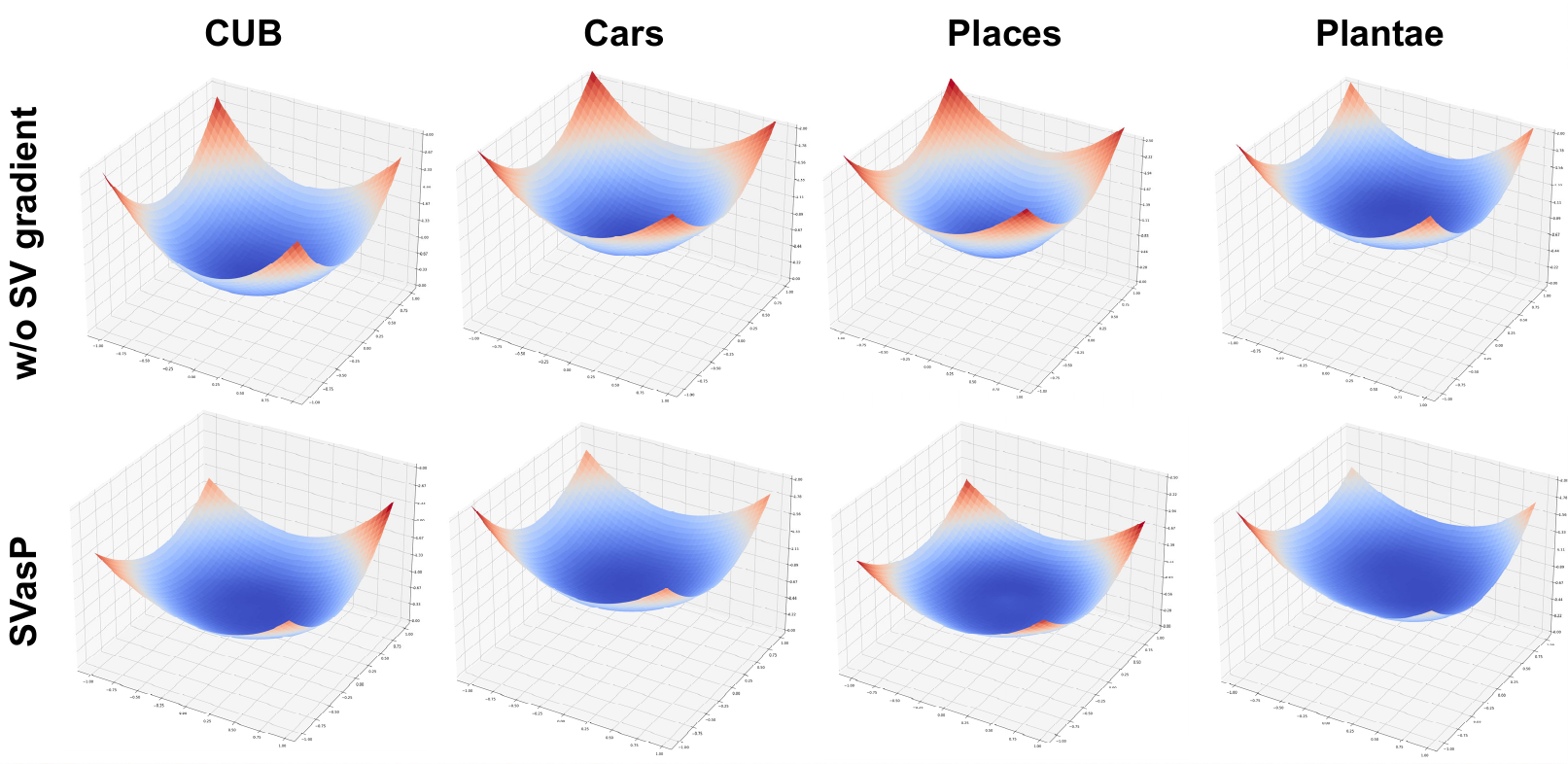} 
\caption{Loss landscape visualization results of the model without SV gradient ensemble module (first row) and our SVasP model (secondc row) on the \textit{mini}-CUB benchmark.} 
\label{fig:supp_landscape}
\end{figure}
\begin{figure}[t!]
\centering
\includegraphics[width=\linewidth]{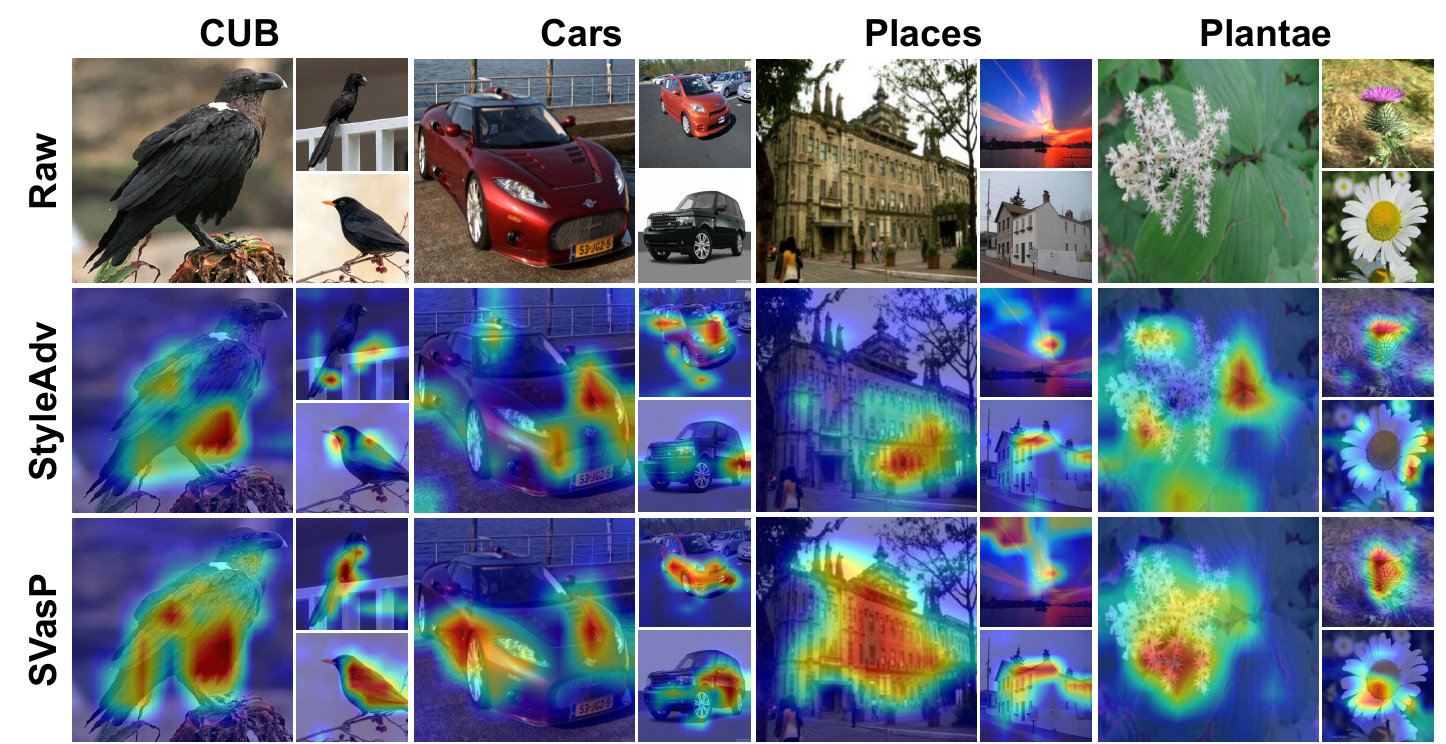} 
\caption{Grad-CAM visualization results of the StyleAdv model and our SVasP model on the \textit{mini}-CUB benchmark. For each target dataset, three examples are demonstrated.}
\label{fig:supp_cam}
\end{figure}
\section{B. More Experimental Results}
\begin{table*}[htbp]
  \centering
  \setlength{\tabcolsep}{0.74mm} 
    \begin{tabular}{@{}ll|cccc|cccc|l@{}}
    \toprule
     & \textbf{$\xi$} & \textbf{ChestX} & \textbf{ISIC} & \textbf{EuroSAT} & \textbf{CropDisease} & \textbf{CUB} & \textbf{Cars} & \textbf{Places} & \textbf{Plantae} & \textbf{Aver.} \\
    \midrule
    &0   & 25.33{$\pm$0.36} & 48.92{$\pm$0.55} & 87.17{$\pm$0.53} & 93.36{$\pm$0.41} & 64.84{$\pm$0.70} &49.29{$\pm$0.65} & 76.02{$\pm$0.62} & 57.24{$\pm$0.66}  & 62.77   \\
    & 0.01   & 25.32{$\pm$0.36} & 47.81{$\pm$0.54} & 86.88{$\pm$0.56} & 92.55{$\pm$0.44} & 66.12{$\pm$0.69} & 48.53{$\pm$0.67} & 75.92{$\pm$0.65} & 58.18{$\pm$0.65} & 62.66  \\
    & \cellcolor{gray!15}{{0.1 (\textbf{Ours}) }}  & \cellcolor{gray!15}{\textbf{26.87{$\pm$0.38}}} & \cellcolor{gray!15}{\textbf{51.10{$\pm$0.58}}} & \cellcolor{gray!15}{\textbf{88.72{$\pm$0.52}}} & \cellcolor{gray!15}{\textbf{94.52{$\pm$0.33}}} & \cellcolor{gray!15}{\textbf{68.95{$\pm$0.66}}} & \cellcolor{gray!15}{\textbf{52.13{$\pm$0.66}}} & \cellcolor{gray!15}{\textbf{77.78{$\pm$0.62}}} & \cellcolor{gray!15}{\textbf{60.63{$\pm$0.64}}} & \cellcolor{gray!15}{\textbf{65.09}} \\
    & 1  & 26.65{$\pm$0.39} & 49.62{$\pm$0.55} & 86.17{$\pm$0.53} & 91.07{$\pm$0.47} & 67.13{$\pm$0.66} & 51.30{$\pm$0.69} & {77.22{$\pm$0.62}} & 58.18{$\pm$0.65} & 63.42  \\
    \bottomrule
    \end{tabular}%
  \caption{More specific results on different decay factors for $\xi$. The accuracy (\%) with RseNet-10 and GNN under the 5-way 5-shot setting is reported. ``\textbf{Aver.}'' means ``Average Accuracy'' of the eight datasets. The optimal results are marked in \textbf{bold}.}
  \label{tab:ab-decay}%
\end{table*}
 \begin{table*}[htbp]
  \centering
  \setlength{\tabcolsep}{0.74mm} 
    \begin{tabular}{@{}ll|cccc|cccc|l@{}}
    \toprule
     & \textbf{$\lambda$} & \textbf{ChestX} & \textbf{ISIC} & \textbf{EuroSAT} & \textbf{CropDisease} & \textbf{CUB} & \textbf{Cars} & \textbf{Places} & \textbf{Plantae} & \textbf{Aver.} \\
    \midrule
    &0   & 26.40{$\pm$0.38} & 49.62{$\pm$0.55} & 87.31{$\pm$0.53} & 93.47{$\pm$0.42} & 67.86{$\pm$0.69} &51.30{$\pm$0.69} & 76.89{$\pm$0.62} & 57.48{$\pm$0.65}  & 63.79  \\
    & \cellcolor{gray!15}{{0.2 (\textbf{Ours})}}  & \cellcolor{gray!15}{\textbf{26.87{$\pm$0.38}}} & \cellcolor{gray!15}{\textbf{51.10{$\pm$0.58}}} & \cellcolor{gray!15}{\textbf{88.72{$\pm$0.52}}} & \cellcolor{gray!15}{\textbf{94.52{$\pm$0.33}}} & \cellcolor{gray!15}{\textbf{68.95{$\pm$0.66}}} & \cellcolor{gray!15}{\textbf{52.13{$\pm$0.66}}} & \cellcolor{gray!15}{77.78{$\pm$0.62}} & \cellcolor{gray!15}{\textbf{60.63{$\pm$0.64}}} & \cellcolor{gray!15}{\textbf{65.09}} \\
    & 0.5   & 26.09{$\pm$0.36} & 49.23{$\pm$0.53} & 88.03{$\pm$0.52} & 92.67{$\pm$0.42} & 67.91{$\pm$0.67} & 50.07{$\pm$0.66} & 76.95{$\pm$0.62} & 58.54{$\pm$0.67} & 63.69  \\
    & 0.8  & 25.61{$\pm$0.37} & 49.03{$\pm$0.53} & 87.35{$\pm$0.51} & 93.34{$\pm$0.40} & 65.53{$\pm$0.68} & 49.62{$\pm$0.63} & 75.58{$\pm$0.64} & 58.32{$\pm$0.63} & 62.72  \\
    & 1  & 26.18{$\pm$0.38} & 49.46{$\pm$0.54} & 87.05{$\pm$0.55} & 92.46{$\pm$0.43} & 67.26{$\pm$0.67} & 50.47{$\pm$0.68} & \textbf{77.84{$\pm$0.61}} & 59.40{$\pm$0.66} & 63.67  \\
    \bottomrule
    \end{tabular}%
  \caption{More specific results on different crop consistency loss functions with different strategies for $\lambda$. The accuracy (\%) with RseNet-10 and GNN under the 5-way 5-shot setting is reported. ``\textbf{Aver.}'' means ``Average Accuracy'' of the eight datasets. The optimal results are marked in \textbf{bold}.}
  \label{tab:ab-lambda}%
\end{table*}%
\begin{table*}[htbp]
  \centering
  \setlength{\tabcolsep}{0.74mm} 
    \begin{tabular}{c|cccc|cccc|c}
    \toprule
    $\kappa_1$$\kappa_2$ & \textbf{ChestX} & \textbf{ISIC} & \textbf{EuroSAT} & \textbf{CropDisease} & \textbf{CUB} & \textbf{Cars} & \textbf{Places} & \textbf{plantae} & \textbf{Average} \\
    \midrule
    Baseline & 26.21±0.31 & 49.99±0.61 & 88.23±0.53 & 93.89±0.32 & 66.49±0.69 & 51.91±0.68 & 75.01±0.64 & 59.20±0.65 & 63.87  \\
    \cellcolor{gray!15}{{\textbf{Ours}}}  & \cellcolor{gray!15}{\textbf{26.87{$\pm$0.38}}} & \cellcolor{gray!15}{\textbf{51.10{$\pm$0.58}}} & \cellcolor{gray!15}{\textbf{88.72{$\pm$0.52}}} & \cellcolor{gray!15}{\textbf{94.52{$\pm$0.33}}} & \cellcolor{gray!15}{\textbf{68.95{$\pm$0.66}}} & \cellcolor{gray!15}{\textbf{52.13{$\pm$0.66}}} & \cellcolor{gray!15}{77.78{$\pm$0.62}} & \cellcolor{gray!15}{\textbf{60.63{$\pm$0.64}}} & \cellcolor{gray!15}{\textbf{65.09}} \\
    \bottomrule
    \end{tabular}%
  \caption{More specific results on different selection methods for $\kappa_1$ and $\kappa_2$. The accuracy (\%) with RseNet-10 and GNN under the 5-way 5-shot setting is reported. ``\textbf{Aver.}'' means ``Average Accuracy'' of the eight datasets.}
  \label{tab:k1_k2}%
\end{table*}%
\begin{table*}[htbp]
  \centering
  \setlength{\tabcolsep}{1.2mm} 
    \begin{tabular}{@{}l|cccc|cccc}
    \toprule
    \textbf{k} & \textbf{ChestX} & \textbf{ISIC} & \textbf{EuroSAT} & \textbf{CropDisease} & \textbf{CUB} & \textbf{Cars} & \textbf{Places} & \textbf{plantae} \\
    \midrule
    0  & 26.23±0.37 & 47.34±0.55 & 86.38±0.54 & 91.71±0.44 & 64.98±0.67 & 47.78±0.64 & 75.58±0.61 & 58.27±0.66 \\
    1  & 25.31±0.37 & 49.26±0.53 & 87.17±0.53 & 93.36±0.41 & 66.18±0.67 & 49.99±0.65 & 76.02±0.62 & 57.32±0.64 \\
    \cellcolor{gray!15}{{2 (\textbf{Ours})}}  & \cellcolor{gray!15}{\textbf{26.87{$\pm$0.38}}} & \cellcolor{gray!15}{\textbf{51.10{$\pm$0.58}}} & \cellcolor{gray!15}{\textbf{88.72{$\pm$0.52}}} & \cellcolor{gray!15}{\textbf{94.52{$\pm$0.33}}} & \cellcolor{gray!15}{\textbf{68.95{$\pm$0.66}}} & \cellcolor{gray!15}{\textbf{52.13{$\pm$0.66}}} & \cellcolor{gray!15}{\textbf{77.78{$\pm$0.62}}} & \cellcolor{gray!15}{\textbf{60.63{$\pm$0.64}}} \\
    3  & 25.46±0.35 & 47.69±0.53 & 88.03±0.52 & 92.99±0.42 & 65.72±0.70 & 50.39±0.66 & 76.85±0.61 & 58.90±0.66 \\
    4  & 25.59±0.37 & 49.59±0.53 & 87.92±0.50 & 94.01±0.38 & 64.84±0.66 & 50.16±0.67 & 75.31±0.64 & 57.66±0.65 \\
    5  & 24.94±0.35 & 48.21±0.52 & 86.54±0.51 & 92.19±0.44 & 62.76±0.69 & 49.17±0.62 & 74.88±0.63 & 55.00±0.63 \\
    \bottomrule
    \end{tabular}%
  \caption{More specific results on crop numbers $k$. The accuracy (\%) with RseNet-10 and GNN under the 5-way 5-shot setting is reported. ``\textbf{Aver.}'' means ``Average Accuracy'' of the eight datasets.}
  \label{tab:k}%
\end{table*}%
\subsubsection{B.1. Impact of different crop numbers $k$.} \leavevmode  \\
The results of the experiments on each dataset for different number of crops $k$ are shown in Table \ref{tab:k}. From the experimental results on each dataset, it can be seen that the optimal classification accuracy is achieved regardless of any dataset when $k=2$. Setting $k$ to 2 is the optimal solution, which can effectively improve the generalization of the model and will not lead to model overfitting.
\subsubsection{B.2. Impact of different strategies for $\xi$ and $\lambda$.}  \leavevmode  \\
The results of the experiments on each dataset for different values of $\xi$ are shown in Table \ref{tab:ab-decay}. As can be seen from the table, when $\xi = 0.1$, a significant performance improvement is realized compared to the other set values, regardless of the dataset. In addition, the results of the experiments for different values of $\lambda$ are shown in Table \ref{tab:ab-lambda}. The optimal classification accuracy is achieved only on the Places dataset by setting the value of $\lambda$ to 1. On the rest of the datasets, optimal classification accuracy is achieved when the value of $\lambda$ is set to 0.2.
\subsubsection{B.3. Impact of different selection methods for $\kappa_1$ and $\kappa_2$.}  \leavevmode  \\
The results of the experiments on each dataset for whether setting $\kappa_1$ and $\kappa_2$ to the same value are shown in Table \ref{tab:k1_k2}. From the results in the table, it can be seen that not restricting $\kappa_1$ and $\kappa_2$ to the same value can effectively increase the accuracy. This is because varying $\kappa_1$ and $\kappa_2$ by different magnitudes can simulate a wider variety of styles, enhancing generalization.
\section{C. More Visualization Results}
\subsubsection{C.1. Loss Landscape Visualization Results.} \leavevmode  \\
We complement the loss landscape visualization results on the \textit{mini}-CUB benchmark, as shown in Figure \ref{fig:supp_landscape}. Consistent with experimental results on the BSCD-FSL benchmark, our proposed method SVasP achieves flatter loss landscape near the optimum that our model converges on.
\subsubsection{C.2. Grad-CAM Visualization Results.} \leavevmode  \\
We also complement Grad-CAM visualization results on the \textit{mini}-CUB benchmark, as shown in Figure \ref{fig:supp_cam}.

\end{document}